\documentclass[journal]{IEEEtran}

\usepackage{cite}
\usepackage{amsmath,amssymb,amsfonts}
\usepackage{graphicx}
\usepackage{subcaption}
\usepackage{booktabs}
\usepackage{multirow}
\usepackage{algorithm}
\usepackage{algorithmic}
\usepackage{array}
\usepackage{url}
\usepackage{xcolor}

\usepackage{orcidlink}
\hypersetup{hidelinks}

\title{Cross-Modal Registration Between 3D and 2D Fingerprints via Pose-Aware Unwrapping and Point-Cloud Fusion}

\author{Xiongjun~Guan$^{\orcidlink{0000-0001-8887-3735}}$,
Jianjiang~Feng$^{\orcidlink{0000-0003-4971-6707}}$,~\IEEEmembership{Member,~IEEE},
and Jie~Zhou$^{\orcidlink{0000-0001-7701-234X}}$,~\IEEEmembership{Senior Member,~IEEE}%
\thanks{Xiongjun Guan, Jianjiang Feng, and Jie Zhou are with the Department of Automation, Tsinghua University, Beijing 100084, China (e-mail: \url{gxj21@mails.tsinghua.edu.cn}; \url{jfeng@tsinghua.edu.cn}; \url{jzhou@tsinghua.edu.cn}).}}

\begin{document}

\maketitle

\begin{abstract}
Three-dimensional (3D) fingerprints preserve global finger geometry and local ridge structure while avoiding contact-induced deformation, but they remain difficult to integrate with legacy two-dimensional (2D) fingerprint systems. This paper addresses the intermediate stage between 3D acquisition and cross-modal matching, and presents a unified framework for 3D fingerprint preprocessing and registration across contactless and contact-based 2D modalities. The framework combines four components: 1) a nonparametric visualization and unwrapping method that converts a 3D fingerprint point cloud into a rolled-equivalent 2D representation without relying on a global finger-shape model; 2) a point-cloud fusion pipeline that registers and mosaics multiple partial 3D captures into a more complete fingerprint model; 3) an ellipse-based pose normalization method for canonical finger alignment; and 4) a pose-aware cross-modal registration strategy that improves compatibility between 3D fingerprints and both contactless and contact-based 2D fingerprints. Experiments on a self-collected multimodal fingerprint database containing 150 fingers show that the proposed framework achieves ridge-level 3D registration accuracy, robust pose estimation, and consistent gains in 2D compatibility. In particular, the 3D fusion error is concentrated around 0.09 mm, contactless 2D--3D registration reaches ridge-scale projection accuracy, and pose-aware unwrapping improves genuine matching scores relative to generic 3D unwrapping. These results support the use of 3D fingerprints as an effective geometric bridge across heterogeneous fingerprint modalities.
The baseline implementation has been publicly released at \url{https://github.com/XiongjunGuan/3DFpVisual}.
\end{abstract}


\section{Introduction}
Fingerprint recognition remains one of the most widely deployed biometric technologies in civilian authentication, access control, mobile devices, and forensic investigation because fingerprints are highly distinctive, persistent over time, and relatively inexpensive to acquire \cite{maltoni2009handbook}. Nevertheless, most operational systems still rely on contact-based 2D sensing. Although such sensors provide clear ridge-valley contrast, they also suffer from limited contact area, sensitivity to skin moisture and sensor contamination, and deformation caused by pressing the finger against the sensing surface.

Compared with contact-based 2D fingerprints, 3D fingerprints acquired without physical contact provide two important advantages. First, they preserve the geometric finger shape together with ridge texture, thereby reducing contact-induced elastic distortion. Second, they offer a natural geometric bridge across heterogeneous acquisition modalities, including contactless 2D images and legacy contact-based 2D fingerprints. In practice, however, the value of 3D fingerprints is still limited by incomplete scans, pose inconsistency across acquisitions, cumbersome acquisition procedures, and the lack of a practical registration mechanism that connects 3D data to mature 2D matching pipelines.

This work addresses the stage between 3D fingerprint acquisition and final identity matching. Rather than replacing mature 2D matchers, we seek to transform 3D fingerprints into representations that are more compatible with existing 2D software while retaining the geometric information available only in 3D. Under this view, 3D fingerprints serve primarily as a gallery-side geometric prior, whereas contactless and contact-based 2D fingerprints remain convenient query-side observations. The resulting objective is to improve inter-session pose consistency, reduce cross-modal deformation, and make a 3D gallery interoperable with both contactless and contact-based queries.

The main contributions of this paper are summarized as follows.
\begin{itemize}
\item We propose a nonparametric 3D fingerprint visualization and unwrapping method that estimates local surface depth directly from point neighborhoods and avoids bias from oversimplified global shape fitting.
\item We develop a 3D fingerprint fusion pipeline that combines minutia-guided coarse registration, rigid refinement, and seam-aware mosaicking to synthesize more complete 3D fingerprints from partial scans.
\item We introduce an ellipse-based pose normalization scheme that estimates a canonical finger pose from cross-sectional geometry and improves inter-session consistency for downstream registration.
\item We present a pose-aware cross-modal registration strategy that improves the compatibility of 3D fingerprints with both contactless and contact-based 2D fingerprints while remaining compatible with legacy 2D matching software.
\end{itemize}

\section{Related Work}
\subsection{3D Fingerprint Unwrapping}
Early 3D-to-2D fingerprint conversion methods usually relied on parametric finger models such as cylinders, spheres, or deformable cylindrical surfaces \cite{chen20063d,zhao20113d,wang2010fit,labati2011fast,labati2012quality,anitha2014performance,dighade2012approach}. Their main advantage is efficiency, but the assumed global shape often mismatches real finger geometry and therefore introduces additional distortion during unfolding. Nonparametric methods reduce this bias by unfolding the surface according to local geometry rather than an explicit global template \cite{fatehpuria2006acquiring,shafaei2009new}. More recently, the literature has expanded toward learning-based reconstruction and pose-aware unwrapping, including monocular 3D reconstruction with subsequent unwarping \cite{cui2023monocular3d} and explicit pose-specific unfolding for contact-based matching \cite{guan2021posespecific}. Our work follows the same motivation of reducing unwrapping-induced mismatch, but emphasizes a broader preprocessing pipeline in which unwrapping, fusion, and pose normalization are jointly designed for cross-modal registration.

\subsection{Distortion Handling in 2D Fingerprints}
Distortion-tolerant fingerprint recognition in the 2D domain has been studied extensively. Existing strategies include distortion rectification before matching \cite{watson2000distortion,si2015detection,dabouei2018dcnn,dabouei2019deep}, global alignment with tolerance margins \cite{jea2005minutia,chen2006new,zheng2007robust,tong2008local}, elastic registration based on thin-plate splines, phase demodulation, or dense deformation estimation \cite{bazen2003fingerprint,ross2006image,si2017dense,cui20182,cui2019drrn,cui2020dense}, and local structural constraints or descriptor learning \cite{kovacs2000fingerprint,cappelli2010minutia,chen2006algorithm,feng2008combining,cheng2013minutiae}. More recent learning-based work has further strengthened this direction, from direct distortion-field regression \cite{guan2022direct} to dense deformation estimation from a single fingerprint image \cite{guan2023distortion}, densely sampled-point matching in difficult cases \cite{gu2021latentreg}, and efficient phase-aware dense registration \cite{guan2024pdrnet}. These methods are highly effective within planar fingerprint domains, but they do not explicitly leverage 3D finger geometry as a bridge across sensing modalities. This difference is central to our formulation.

\subsection{Fingerprint Pose Estimation and Contactless Alignment}
Pose estimation has become increasingly important in fingerprint sensing scenarios where geometric inconsistency, view angle, and sensor-specific observation conditions substantially affect matching accuracy. This trend appears in several settings, including fingerprint-image-based finger-angle estimation \cite{he2022fingerangle}, 2D--3D fingerprint matching for finger-pose recovery \cite{duan2022pose3d}, dense-voting-based pose estimation from fingerprints \cite{duan2023densevoting}, and under-screen fingerprint sensing where explicit pose estimation compensates for the mismatch between sensing geometry and the effective observation model \cite{guan2025underscreenpose}. In parallel, contactless recognition has become increasingly sensitive to image quality and viewpoint variation \cite{priesnitz2024mclfiq}, which has motivated 3D-pose-guided contactless recognition pipelines \cite{pei2025contactless3dpose}. These works reinforce the importance of pose as a first-class factor in fingerprint interoperability. Our setting differs in that pose estimation is used not as an endpoint, but as an intermediate step for cross-modal re-unwrapping and registration.

\subsection{3D Fingerprint Acquisition and Reconstruction}
Several acquisition paradigms have been proposed for touchless 3D fingerprints, including multi-camera systems, structured light, photometric reconstruction, direct 3D sensing, and multibiometric 3D finger capture \cite{parziale2006surround,wang2010data,kumar20183d,galbally2017full,arora20143dphantoms,yang2022multibiometrics}. Some methods reconstruct 3D fingerprints from multiple contactless 2D images \cite{liu20143d,labati2015toward,kumar2013towards}, whereas others recover richer 3D structure or even 3D minutiae from a single or small number of contactless views \cite{cui2023monocular3d,dong2025bridging3dminutiae}. At the representation level, recent work has also explored dense minutia descriptors and fixed-length dense fingerprint representations for modern matching pipelines \cite{pan2024denseminutia,pan2024fdd,pan2025fixedlength}. In contrast, our goal is not to redesign the entire recognition stack, but to make already-acquired 3D point clouds more compatible with legacy multi-modal fingerprint systems through geometry-aware preprocessing and registration.

\begin{figure*}[!t]
	\centering
	\includegraphics[width=0.95\linewidth]{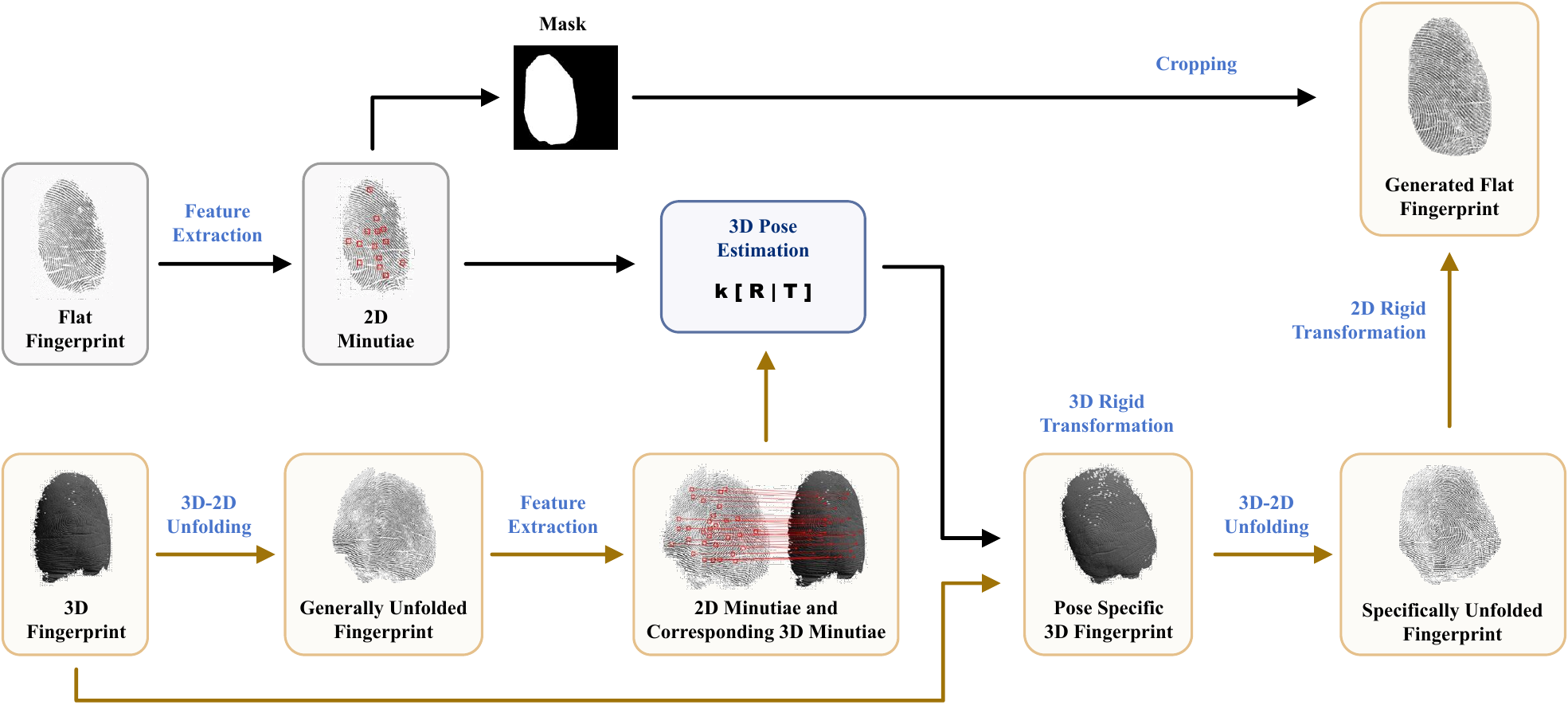}
	\caption{Overall workflow of the proposed framework.}
	\label{fig:framework}
\end{figure*}

\section{Proposed Method}
\subsection{Problem Formulation and Pipeline Overview}
We consider three fingerprint modalities collected from the same finger: a 3D fingerprint point cloud, one or more contactless 2D fingerprint images, and one or more contact-based 2D fingerprints. Let the 3D fingerprint be represented as a point set
\begin{equation}
\mathbf{V}=\{\mathbf{v}_i\in\mathbb{R}^3\}_{i=1}^{N}.
\end{equation}
Our goal is to transform $\mathbf{V}$ into representations that can be aligned with both types of 2D fingerprints while preserving ridge-level geometry and reducing cross-modal deformation.

The proposed pipeline is illustrated in Fig.~\ref{fig:framework}. Starting from one or more point clouds, we first visualize the 3D surface by estimating point-wise local depth. The resulting image is then nonparametrically unwrapped into a rolled-equivalent 2D representation. If multiple partial 3D captures are available, we register and mosaic them into a more complete point cloud. The fused point cloud is normalized to a canonical pose using cross-sectional ellipse fitting. Finally, we register the normalized 3D fingerprint with contactless 2D fingerprints via camera-model-based 2D--3D alignment, and with contact-based 2D fingerprints via pose-aware re-unwrapping.

\subsection{3D Fingerprint Visualization}
Direct rendering of a 3D fingerprint with conventional illumination often yields unstable ridge contrast because the observed brightness depends on lighting and view direction. To obtain a 2D representation more consistent with the ridge-valley topology of a contact fingerprint, we estimate the local surface depth of each point from its neighborhood.

For any point $\mathbf{p}\in\mathbf{V}$, its neighborhood is defined as
\begin{equation}
\mathbf{X}=\{\mathbf{x}_i\mid \mathbf{x}_i\in\mathbf{V},\ \|\mathbf{x}_i-\mathbf{p}\|<r\},
\label{eq:def_X}
\end{equation}
where $r$ is the neighborhood radius. The neighborhood centroid is
\begin{equation}
\mathbf{c}=\operatorname{MEAN}(\mathbf{X}).
\label{eq:def_c}
\end{equation}
The local normal vector $\mathbf{n}$ is estimated by minimizing the squared orthogonal distances from the neighborhood points to the local tangent plane:
\begin{equation}
\mathbf{n}=
\arg\min_{\|\mathbf{n}\|=1}
\sum_{\mathbf{x}_i\in\mathbf{X}}
\left((\mathbf{x}_i-\mathbf{c})^{T}\mathbf{n}\right)^2.
\label{eq:def_n}
\end{equation}
This optimization can be solved by principal component analysis, where $\mathbf{n}$ is the eigenvector corresponding to the smallest eigenvalue of the local covariance matrix.

The local surface depth at $\mathbf{p}$ is then defined by projecting the offset $(\mathbf{p}-\mathbf{c})$ onto the local normal:
\begin{equation}
d=(\mathbf{p}-\mathbf{c})^{T}\mathbf{n}.
\label{eq:def_d}
\end{equation}
After computing the depth values for all observable points, the values are normalized into the range $[0,1]$:
\begin{equation}
d_n=\frac{d-d_{\min}}{d_{\max}-d_{\min}},
\label{eq:def_dn}
\end{equation}
where $d_{\max}$ and $d_{\min}$ denote the global maximum and minimum surface depths in the current point cloud. Before visualization, we further discard invisible points whose local normals form an angle larger than $85^\circ$ with the normal of the viewing plane. The normalized depth is then used as the grayscale intensity. Since sparse sampling may leave holes after orthographic projection, we fill the missing areas by interpolation and apply histogram equalization to enhance ridge contrast in the final visualization.

\begin{figure}[!t]
\centering
\includegraphics[width=0.9\linewidth]{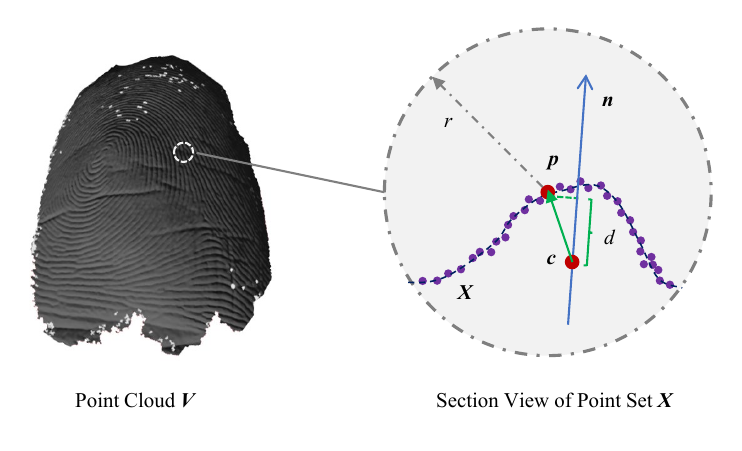}
\caption{Illustration of the local surface-depth visualization strategy.}
\label{fig:visual_method}
\end{figure}

This construction has two advantages. First, the generated grayscale is tied to local geometry rather than global illumination. Second, the polarity of ridges and valleys remains consistent across the image, making the result more compatible with downstream fingerprint feature extraction.

\begin{figure}[!t]
\centering
\subcaptionbox{Thumb}{\includegraphics[width=0.32\linewidth]{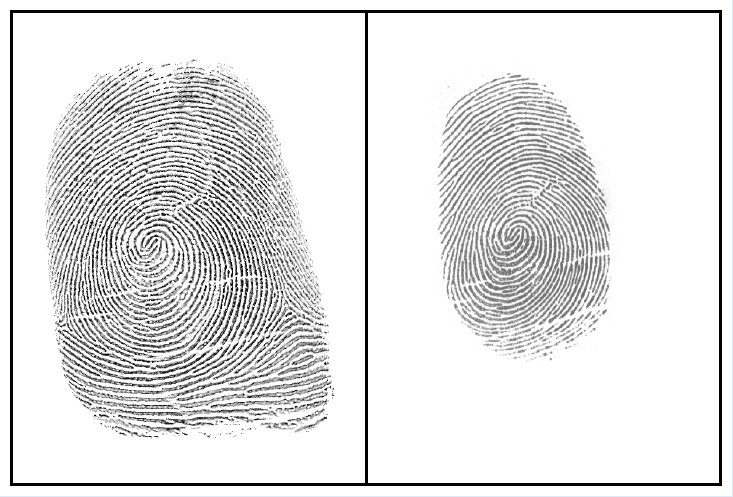}}
\hfill
\subcaptionbox{Middle}{\includegraphics[width=0.32\linewidth]{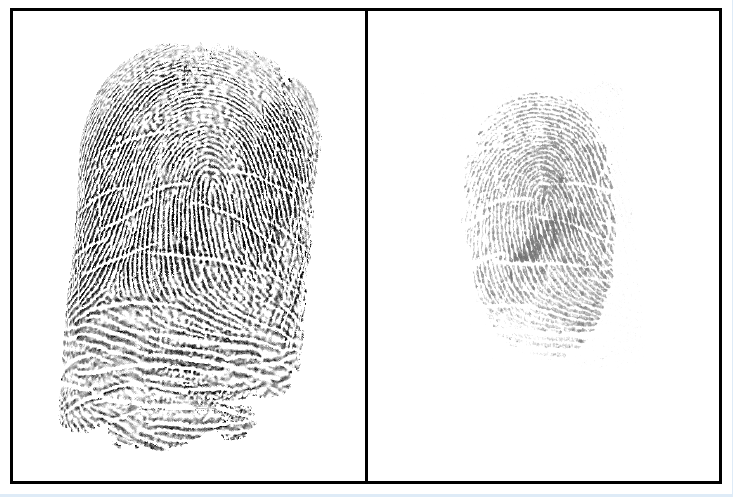}}
\hfill
\subcaptionbox{Little}{\includegraphics[width=0.32\linewidth]{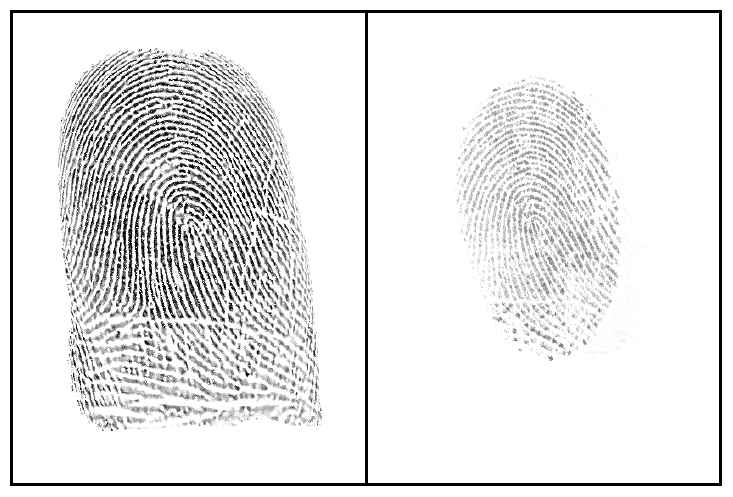}}
\caption{Examples of 3D fingerprint visualization results.}
\label{fig:visual_examples}
\end{figure}

\begin{algorithm}[!t]
\caption{3D fingerprint visualization by local surface depth}
\label{alg:visual}
\begin{algorithmic}[1]
\REQUIRE Point cloud $\mathbf{V}$, neighborhood radius $r$
\ENSURE Gray-value set $B$ and normal set $\mathbf{N}$
\STATE Initialize $D\leftarrow \emptyset$, $\mathbf{N}\leftarrow \emptyset$
\FORALL{$\mathbf{p}\in\mathbf{V}$}
\STATE Construct neighborhood $\mathbf{X}=\{\mathbf{x}_i\in\mathbf{V}\mid \|\mathbf{x}_i-\mathbf{p}\|<r\}$
\STATE Compute centroid $\mathbf{c}=\operatorname{MEAN}(\mathbf{X})$
\STATE Estimate local normal $\mathbf{n}$ by PCA on $\mathbf{X}$
\IF{$(\mathbf{p}-\mathbf{c})^T\mathbf{n}<0$}
\STATE $\mathbf{n}\leftarrow -\mathbf{n}$
\ENDIF
\STATE $d\leftarrow (\mathbf{p}-\mathbf{c})^T\mathbf{n}$
\STATE Append $d$ to $D$ and $\mathbf{n}$ to $\mathbf{N}$
\ENDFOR
\STATE Normalize $D$ into $B=(D-\min D)/(\max D-\min D)$
\RETURN $B,\mathbf{N}$
\end{algorithmic}
\end{algorithm}

\subsection{Nonparametric Unwrapping}
The orthographic visualization still represents a curved finger surface and is therefore not yet equivalent to a contact-based rolled fingerprint. To reduce this discrepancy, we unfold the visualization according to local arc length rather than fitting a global finger-shape model.

Let $z(x,y)$ denote the depth map induced by the projected point cloud. The gradients along the two image axes are
\begin{equation}
(g_x,g_y)=\left(\frac{\partial z}{\partial x},\frac{\partial z}{\partial y}\right).
\label{eq:grad}
\end{equation}
The local relation between the original image coordinates $(x,y)$ and the unwrapped coordinates $(u,v)$ is defined by arc-length accumulation:
\begin{equation}
\frac{\partial u}{\partial x}=\sqrt{1+\left(\frac{\partial z}{\partial x}\right)^2},
\qquad
\frac{\partial u}{\partial y}=\sqrt{1+\left(\frac{\partial z}{\partial y}\right)^2},
\label{eq:uv_u}
\end{equation}
and similarly
\begin{equation}
\frac{\partial v}{\partial x}=\sqrt{1+\left(\frac{\partial z}{\partial x}\right)^2},
\qquad
\frac{\partial v}{\partial y}=\sqrt{1+\left(\frac{\partial z}{\partial y}\right)^2}.
\label{eq:uv_v}
\end{equation}
In practice, we choose the image location with the smallest gradient magnitude as the unfolding center $(c_x,c_y)$ and compute
\begin{equation}
u(x,y)=\int_{c_x}^{x}\sqrt{1+g_x(\xi,y)^2}\,d\xi,
\label{eq:u_int}
\end{equation}
\begin{equation}
v(x,y)=\int_{c_y}^{y}\sqrt{1+g_y(x,\eta)^2}\,d\eta.
\label{eq:v_int}
\end{equation}
The original visualization is then remapped according to
\begin{equation}
(x,y)\mapsto (u+c_x,v+c_y).
\label{eq:map_unwrap}
\end{equation}

Compared with a global cylindrical model, this strategy preserves local geometry more faithfully and reduces over-smoothing of irregular finger shape. It is particularly useful when the side regions of the finger are strongly curved relative to the viewing plane.

\begin{figure}[!t]
\centering
\subcaptionbox{Frontal view}{\includegraphics[width=0.45\linewidth]{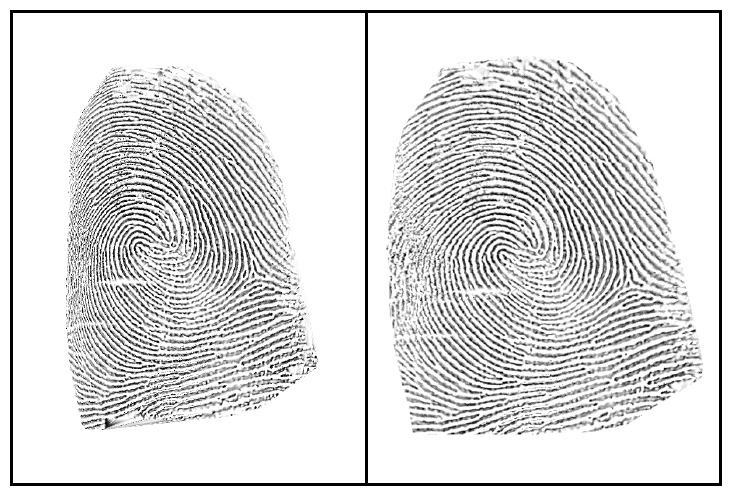}}
\hfill
\subcaptionbox{Side view}{\includegraphics[width=0.45\linewidth]{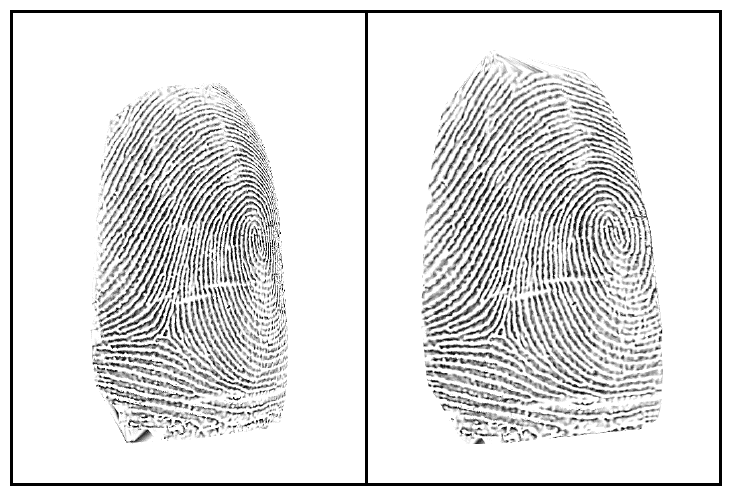}}
\caption{Visualization before and after nonparametric unwrapping.}
\label{fig:unfold}
\end{figure}

\begin{algorithm}[!t]
\caption{Nonparametric unwrapping by local arc-length integration}
\label{alg:unfold}
\begin{algorithmic}[1]
\REQUIRE Visualized image $I_{in}(x,y)$ and gradients $g_x,g_y$
\ENSURE Unwrapped image $I_{out}(x,y)$
\STATE Find the image point $(c_x,c_y)$ with the smallest gradient magnitude
\FORALL{$x$}
\STATE Compute $u(x,y)=\int_{c_x}^{x}\sqrt{1+g_x(\xi,y)^2}\,d\xi$
\ENDFOR
\FORALL{$y$}
\STATE Compute $v(x,y)=\int_{c_y}^{y}\sqrt{1+g_y(x,\eta)^2}\,d\eta$
\ENDFOR
\STATE Remap $I_{in}$ according to $(x,y)\mapsto (u+c_x,v+c_y)$
\RETURN $I_{out}$
\end{algorithmic}
\end{algorithm}

\subsection{3D Fingerprint Fusion}
\subsubsection{Minutia Correspondence Extraction}
Incomplete 3D captures are common in practice, especially when a commodity scanner observes only one side of the finger at a time. To construct a more complete 3D fingerprint, we fuse two or more partial point clouds of the same finger.

We first generate unwrapped 2D visualizations for each point cloud and extract minutiae from them using a commercial fingerprint software package \cite{verifinger}. Minutia correspondences are established with Minutia Cylinder-Code descriptors \cite{cappelli2010minutia} and a spectral matching method \cite{leordeanu2005spectral}. Since each 2D minutia is linked to a known 3D point in the original point cloud, the 2D correspondences induce 3D matched point sets
\begin{equation}
\mathbf{P}=\{\mathbf{p}_1,\mathbf{p}_2,\ldots,\mathbf{p}_n\}, \qquad
\mathbf{Q}=\{\mathbf{q}_1,\mathbf{q}_2,\ldots,\mathbf{q}_n\}.
\end{equation}
This bridge from 2D minutia correspondences to 3D point correspondences is important in practice: it allows the fusion pipeline to operate directly on point clouds while still benefiting from the maturity of existing 2D fingerprint feature extractors.

\subsubsection{Coarse Rigid Registration}
Since the finger is not in contact with any deforming surface during 3D acquisition, we model the transformation between two partial scans by a rigid motion. Let the rotation and translation be $\mathbf{R}$ and $\mathbf{t}$. The registration problem is formulated as
\begin{equation}
\mathbf{R},\mathbf{t}
=
\arg\min_{\mathbf{R},\mathbf{t}}
\sum_{i=1}^{n}
w_i\left\|(\mathbf{R}\mathbf{p}_i+\mathbf{t})-\mathbf{q}_i\right\|^2,
\label{eq:Rt}
\end{equation}
where $w_i$ is the weight of the $i$th correspondence. In our implementation, all matched minutiae are assigned equal weights.

For a fixed rotation $\mathbf{R}$, differentiating the objective with respect to $\mathbf{t}$ yields
\begin{equation}
\mathbf{t}=\bar{\mathbf{q}}-\mathbf{R}\bar{\mathbf{p}},
\label{eq:t}
\end{equation}
where
\begin{equation}
\bar{\mathbf{p}}=\frac{1}{n}\sum_{i=1}^{n}\mathbf{p}_i,\qquad
\bar{\mathbf{q}}=\frac{1}{n}\sum_{i=1}^{n}\mathbf{q}_i.
\end{equation}
Substituting \eqref{eq:t} into \eqref{eq:Rt} gives an equivalent problem on centered point sets:
\begin{equation}
\min_{\mathbf{R}}
\sum_{i=1}^{n}
\left\|
\mathbf{R}(\mathbf{p}_i-\bar{\mathbf{p}})
-
(\mathbf{q}_i-\bar{\mathbf{q}})
\right\|^2.
\label{eq:R_only}
\end{equation}
Let
\begin{equation}
\mathbf{H}=\sum_{i=1}^{n}(\mathbf{p}_i-\bar{\mathbf{p}})(\mathbf{q}_i-\bar{\mathbf{q}})^T.
\end{equation}
If the singular value decomposition of $\mathbf{H}$ is
\begin{equation}
\mathbf{H}=\mathbf{U}\mathbf{\Sigma}\mathbf{V}^T,
\end{equation}
then the optimal rotation is
\begin{equation}
\mathbf{R}=\mathbf{V}\mathbf{U}^T,
\label{eq:R_svd}
\end{equation}
with the usual determinant correction when $\det(\mathbf{R})<0$. This yields an initial rigid alignment robust enough for subsequent local refinement. The formulation is consistent with classical rigid registration methods for range data \cite{chen1992object,besl1992method}, although the correspondences here are fingerprint-driven rather than nearest-neighbor driven.

\subsubsection{Seam-Aware Mosaicking}
After coarse alignment, we refine the point-cloud overlap by rigid registration and then fuse the point clouds through seam-aware mosaicking. In implementation, this refinement follows the standard rigid point-cloud registration workflow used in practical toolkits \cite{pcregrigid2021}. Let $pos_a(r,c)$ and $pos_b(r,c)$ denote the mean 3D coordinates corresponding to pixel $(r,c)$ in the orthographic projections of two aligned point clouds. The local registration disagreement is defined as
\begin{equation}
L_{dis}(r,c)=\|pos_a(r,c)-pos_b(r,c)\|.
\label{eq:Ldis}
\end{equation}
To prevent the seam from drifting toward unstable overlap boundaries, we introduce a center-seeking regularizer
\begin{equation}
L_{cen}(r,c)=
\exp\left(-k \cdot \min\left(\operatorname{distance}(edge,(r,c))\right)\right),
\label{eq:Lcen}
\end{equation}
where $edge$ is the set of overlap-boundary pixels and $k$ is the projection grid size. The total seam penalty is
\begin{equation}
L(r,c)=
\left\{
\begin{array}{ll}
L_{dis}(r,c)+\lambda L_{cen}(r,c), & (r,c)\in S,\\
\infty, & (r,c)\notin S,
\end{array}
\right.
\label{eq:penalty}
\end{equation}
where $S$ is the overlap region. The seam between two endpoints is obtained as the shortest path on an 8-connected graph whose edge weights are induced by $L$. Points near the seam are blended by distance-based weighting to smooth the final transition.

More specifically, let $V_{left}$ and $V_{right}$ denote the points extracted from a narrow band centered on the seam, and let $E_{left}$ and $E_{right}$ denote the corresponding boundary point sets of the band. For a candidate point $p_c$ in the seam band, its nearest neighbors on both sides are
\begin{equation}
p_L=\arg\min_{p\in V_{left}}\operatorname{distance}(p_c,p),
\end{equation}
\begin{equation}
p_R=\arg\min_{p\in V_{right}}\operatorname{distance}(p_c,p).
\end{equation}
The blending weights are defined by the distances to the opposite boundaries:
\begin{equation}
\begin{aligned}
w_R&=\min_{p\in E_{left}}\operatorname{distance}(p_c,p),\\
w_L&=\min_{p\in E_{right}}\operatorname{distance}(p_c,p),
\end{aligned}
\end{equation}
and the final blended point is
\begin{equation}
p^{*}=\frac{p_L w_L+p_R w_R}{w_L+w_R}.
\label{eq:blend}
\end{equation}
This weighting favors the side that is locally farther from its boundary and therefore more reliable at that seam location.

To determine the seam endpoints, we first extract the outermost 8-connected boundaries of the two projected masks and collect their intersection points as seam-endpoint candidates. If there are more than two candidates, we select the farthest pair. If no intersection point exists, one projected region is treated as being fully contained in the other, and explicit mosaicking is skipped. This rule reduces unnecessary seam construction when one scan already covers the effective foreground of the other.

\begin{figure}[!t]
\centering
\subcaptionbox{$L_{dis}$}{\includegraphics[width=0.3\linewidth]{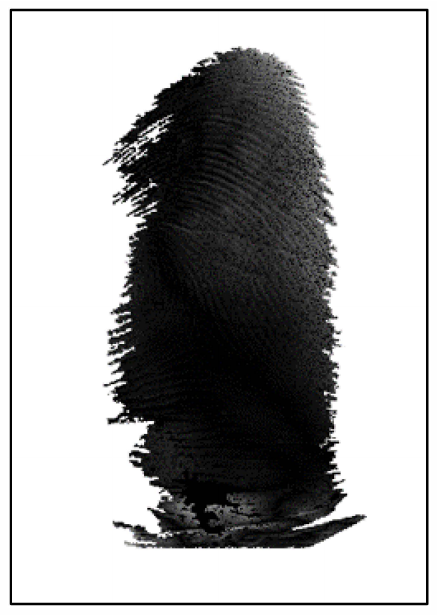}}
\hfill
\subcaptionbox{$L_{cen}$}{\includegraphics[width=0.3\linewidth]{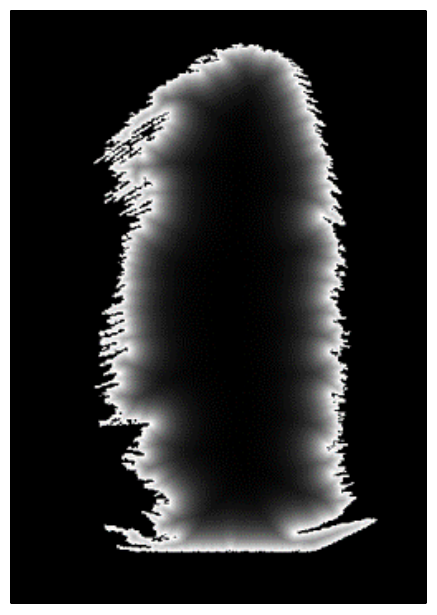}}
\hfill
\subcaptionbox{$L$}{\includegraphics[width=0.3\linewidth]{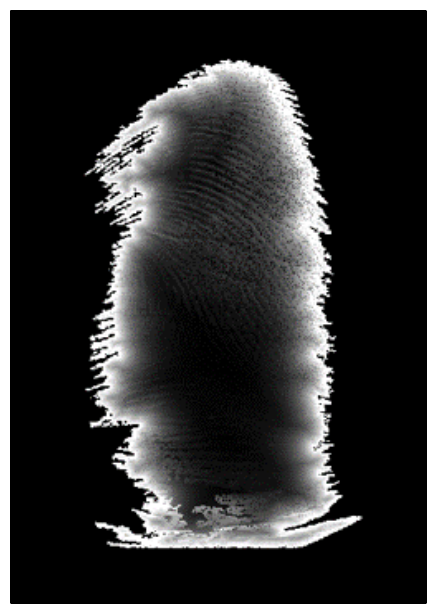}}
\caption{Visualization of seam penalties used in point-cloud mosaicking.}
\label{fig:penalty}
\end{figure}

\begin{algorithm}[!t]
\caption{Seam-aware point-cloud fusion}
\label{alg:point_merge}
\begin{algorithmic}[1]
\REQUIRE Two aligned point clouds $\mathbf{P}_1,\mathbf{P}_2$, penalty weight $\lambda$
\ENSURE Fused point cloud $\mathbf{P}_m$
\STATE Generate orthographic visualizations, coordinate maps, and foreground masks
\STATE Extract boundary intersections and choose the two farthest seam endpoints
\STATE Compute $L_{dis}$ from 3D disagreement and $L_{cen}$ from overlap centrality
\STATE Form the total penalty map $L=L_{dis}+\lambda L_{cen}$
\STATE Find the minimum-cost seam using Dijkstra's algorithm
\STATE Keep the dominant side of each point cloud and blend points near the seam
\STATE Merge all retained and blended points into $\mathbf{P}_m$
\RETURN $\mathbf{P}_m$
\end{algorithmic}
\end{algorithm}

\subsection{Pose Normalization}
The fused point cloud still contains arbitrary pose variation, which is inconvenient for later cross-modal registration. We therefore define a canonical finger pose by estimating the finger center axis and the residual in-plane orientation.

The finger is approximated as a stack of thin elliptical cross-sections. For each slice, the ellipse is parameterized as
\begin{equation}
Ax^2+Bxy+Cy^2+Dx+Ey+F=0,
\label{eq:ellipse}
\end{equation}
subject to the ellipse constraint
\begin{equation}
4AC-B^2>0.
\label{eq:ellipse_constraint}
\end{equation}
Let
\begin{equation}
\mathbf{W}=[A,B,C,D,E,F]^T,
\qquad
\mathbf{X}=[x^2,xy,y^2,x,y,1]^T.
\end{equation}
The optimal ellipse fit is obtained by minimizing
\begin{equation}
\min \ \|\mathbf{W}^T\mathbf{X}\|^2=\mathbf{W}^T\mathbf{X}\mathbf{X}^T\mathbf{W}
\end{equation}
with the quadratic constraint
\begin{equation}
\mathbf{W}^T\mathbf{H}\mathbf{W}=1,
\end{equation}
where
\begin{equation}
\mathbf{H}=
\begin{pmatrix}
0 & 0 & 2 & 0 & 0 & 0\\
0 & -1 & 0 & 0 & 0 & 0\\
2 & 0 & 0 & 0 & 0 & 0\\
0 & 0 & 0 & 0 & 0 & 0\\
0 & 0 & 0 & 0 & 0 & 0\\
0 & 0 & 0 & 0 & 0 & 0
\end{pmatrix}.
\end{equation}
This leads to a generalized eigenvalue problem; the valid solution satisfying the ellipse constraint is selected. The center coordinates of the ellipse are
\begin{equation}
X_C=\frac{BE-2CD}{4AC-B^2},
\qquad
Y_C=\frac{BD-2AE}{4AC-B^2},
\label{eq:ellipse_center}
\end{equation}
and the semi-axis lengths are
\begin{equation}
a=
\sqrt{
\frac{2(AX_C^2+CY_C^2+BX_CY_C-F)}
{A+C+\sqrt{(A-C)^2+B^2}}
},
\end{equation}
\begin{equation}
b=
\sqrt{
\frac{2(AX_C^2+CY_C^2+BX_CY_C-F)}
{A+C-\sqrt{(A-C)^2+B^2}}
}.
\end{equation}
The major-axis orientation is estimated as
\begin{equation}
\theta=\frac{1}{2}\arctan\frac{B}{A-C},
\label{eq:angle}
\end{equation}
with the following case distinctions to resolve axis ordering:
\begin{equation}
\theta=
\left\{
\begin{array}{ll}
0, & B=0 \ \text{and}\ AF<CF,\\
90^\circ, & B=0 \ \text{and}\ AF\geq CF,\\
\frac{1}{2}\operatorname{arccot}\frac{A-C}{B}, & B\neq 0 \ \text{and}\ AF<CF,\\
\frac{1}{2}\arctan\frac{A-C}{B}, & B\neq 0 \ \text{and}\ AF\geq CF.
\end{array}
\right.
\label{eq:angle_cases}
\end{equation}
To reduce the influence of finger tips and joints, only slices within the stable central band of the finger are used in the final line fitting step.
More specifically, we keep slices below the top 20\% of the finger and above the bottom 10\%, so that local irregularities near the finger tip and joint do not dominate the fitted center axis.

By fitting ellipses to multiple slices within the stable central region of the finger and linearly fitting their centers, we obtain the finger center axis. The canonical pose is then defined by aligning this axis with the global vertical direction and resolving the residual roll according to the ellipse orientation.

\subsection{Cross-Modal Registration}
\subsubsection{Registration Between Contactless 2D and 3D Fingerprints}
Because contactless 2D fingerprints and 3D fingerprints are both acquired without pressing the finger against a surface, their geometry differs mainly by camera projection and pose. Therefore, the problem can be modeled as a 2D--3D calibration problem.

Let $(x,y,z,1)^T$ be a 3D minutia in homogeneous coordinates and $(u,v,1)^T$ be its corresponding 2D contactless minutia. Their relationship is
\begin{equation}
\lambda
\begin{bmatrix}
u\\v\\1
\end{bmatrix}
=
\mathbf{K}
\begin{bmatrix}
\mathbf{R} & \mathbf{t}
\end{bmatrix}
\begin{bmatrix}
x\\y\\z\\1
\end{bmatrix},
\label{eq:KRt}
\end{equation}
where $\mathbf{K}$ is the camera intrinsic matrix,
\begin{equation}
\mathbf{K}=
\begin{bmatrix}
f_x & s & c_x\\
0 & f_y & c_y\\
0 & 0 & 1
\end{bmatrix}.
\end{equation}
The projection matrix $\mathbf{P}=\mathbf{K}[\mathbf{R}\mid\mathbf{t}]$ is estimated from matched 2D--3D minutiae by minimizing the geometric reprojection error
\begin{equation}
\begin{aligned}
E_g=\frac{1}{n}\sum_{i=1}^{n}
\Bigg[
&\left(u_i-\frac{\mathbf{p}_1^T(x_i,y_i,z_i)^T+p_{14}}
{\mathbf{p}_3^T(x_i,y_i,z_i)^T+p_{34}}\right)^2 \\
+\ &
\left(v_i-\frac{\mathbf{p}_2^T(x_i,y_i,z_i)^T+p_{24}}
{\mathbf{p}_3^T(x_i,y_i,z_i)^T+p_{34}}\right)^2
\Bigg],
\end{aligned}
\label{eq:Eg}
\end{equation}
where $\mathbf{p}_1^T,\mathbf{p}_2^T,\mathbf{p}_3^T$ are the row vectors of $\mathbf{P}$. Once $\mathbf{P}$ is estimated, it is decomposed into intrinsic and extrinsic parameters. Let
\begin{equation}
\mathbf{P}=\left[\mathbf{H}_{3\times 3},\mathbf{h}_{3\times 1}\right],
\end{equation}
where $\mathbf{H}=\mathbf{K}\mathbf{R}$ and $\mathbf{h}=\mathbf{K}\mathbf{t}$. By applying QR decomposition to $\mathbf{H}^{-1}$,
\begin{equation}
\mathbf{H}^{-1}=\mathbf{Q}_d\mathbf{R}_d,
\end{equation}
we obtain
\begin{equation}
\mathbf{K}=\mathbf{R}_d^{-1},\qquad
\mathbf{R}=\mathbf{Q}_d^T.
\end{equation}
After standard sign correction to ensure positive diagonal entries in $\mathbf{K}$, the translation is recovered from
\begin{equation}
\mathbf{t}=\mathbf{S}\mathbf{R}_d\mathbf{h},
\end{equation}
where $\mathbf{S}$ is the diagonal sign-correction matrix. This decomposition follows standard camera-geometry practice \cite{ganapathy1984decomposition,faugeras1993three} and yields a physically interpretable camera model and finger pose for contactless 2D--3D registration.

\subsubsection{Registration Between Contact-Based 2D and 3D Fingerprints}
The alignment between a 3D fingerprint and a contact-based 2D fingerprint is more difficult because pressing the finger against the sensor introduces non-negligible deformation. Directly unwrapping the 3D point cloud into a generic rolled-equivalent image often leaves substantial pose-induced discrepancy with the target contact fingerprint. To address this issue, we propose pose-aware unwrapping.

Suppose a 3D minutia $(x,y,z)$ corresponds to a contact-based 2D minutia $(r,c)$. We estimate a 3D rigid transformation that best explains the 2D observation under orthographic projection:
\begin{equation}
\begin{bmatrix}
u\\v\\0
\end{bmatrix}
=
\begin{bmatrix}
1&0&0\\
0&1&0\\
0&0&0
\end{bmatrix}
\left(
k
\begin{bmatrix}
\mathbf{R}_3 & \mathbf{t}_3
\end{bmatrix}
\begin{bmatrix}
x\\y\\z\\1
\end{bmatrix}
\right),
\label{eq:pose_contact}
\end{equation}
where $k$ is a scale factor and $(\mathbf{R}_3,\mathbf{t}_3)$ denote the target finger pose under contact conditions. The pose is obtained by minimizing the average 2D projection error:
\begin{equation}
\min \ \frac{1}{n}\sum_{i=1}^{n}\sqrt{(u_i-r_i)^2+(v_i-c_i)^2}.
\label{eq:contact_obj}
\end{equation}

After estimating the target pose, we rotate the 3D point cloud accordingly and perform the unwrapping step again. The second unwrapping is therefore conditioned on the observed contact pose rather than on a generic canonical orientation. A final 2D rigid alignment and cropping step removes residual planar offsets. In practice, this reduces pose-induced deformation before standard 2D matching.

\begin{figure}[!t]
\centering
\subcaptionbox{Contact 2D}{\includegraphics[width=0.3\linewidth]{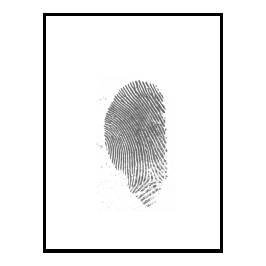}}
\hfill
\subcaptionbox{Raw 3D pose}{\includegraphics[width=0.3\linewidth]{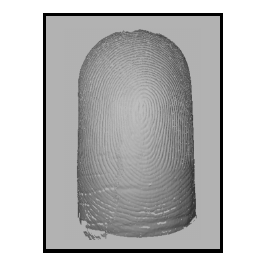}}
\hfill
\subcaptionbox{Estimated pose}{\includegraphics[width=0.3\linewidth]{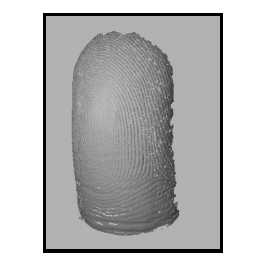}}
\caption{Example of pose estimation for contact-based 2D--3D registration.}
\label{fig:register123}
\end{figure}

\subsection{Method Summary}
The proposed method can be interpreted as a geometry-preserving bridge between heterogeneous fingerprint modalities. The 3D visualization and unwrapping stages convert point-cloud geometry into a ridge-compatible 2D representation. The fusion and pose-normalization stages improve completeness and consistency at the gallery side. The two cross-modal registration branches then adapt the same 3D reference to contactless and contact-based 2D fingerprints, respectively. This organization allows the whole system to reuse mature 2D fingerprint software while still benefiting from 3D geometric information. In summary, the method chapter integrates what were originally separate preprocessing, alignment, and registration blocks into one coherent pipeline: acquire partial 3D geometry, recover a complete and normalized 3D reference, and then use this reference as a modality bridge for both 2D sensing regimes.

\section{Experimental Setup and Results}
\subsection{Dataset and Acquisition Protocol}
All experiments are conducted on a self-collected multimodal fingerprint database because publicly available datasets did not simultaneously satisfy the requirements of this work: 3D fingerprints, contactless 2D images, contact-based 2D fingerprints, and multiple poses for each finger.

The database contains 150 fingers from 15 subjects between 20 and 30 years old. For each finger, two partial 3D scans were collected from different side views with a commercial structured-light scanner, three contactless 2D images were acquired using a smartphone camera, and a temporal sequence of contact-based fingerprints was captured by an FTIR sensor. Eight representative contact-based fingerprints were uniformly sampled from each rolling sequence. The 3D acquisition required approximately 3 min per finger, the contactless 2D acquisition required around 45 s per finger, and the contact-based capture sequence required around 30 s per finger.
The subject pool contains 13 male subjects and 2 female subjects, and all ten fingers of each subject were collected.

\begin{figure}[!t]
\centering
\includegraphics[width=0.95\linewidth]{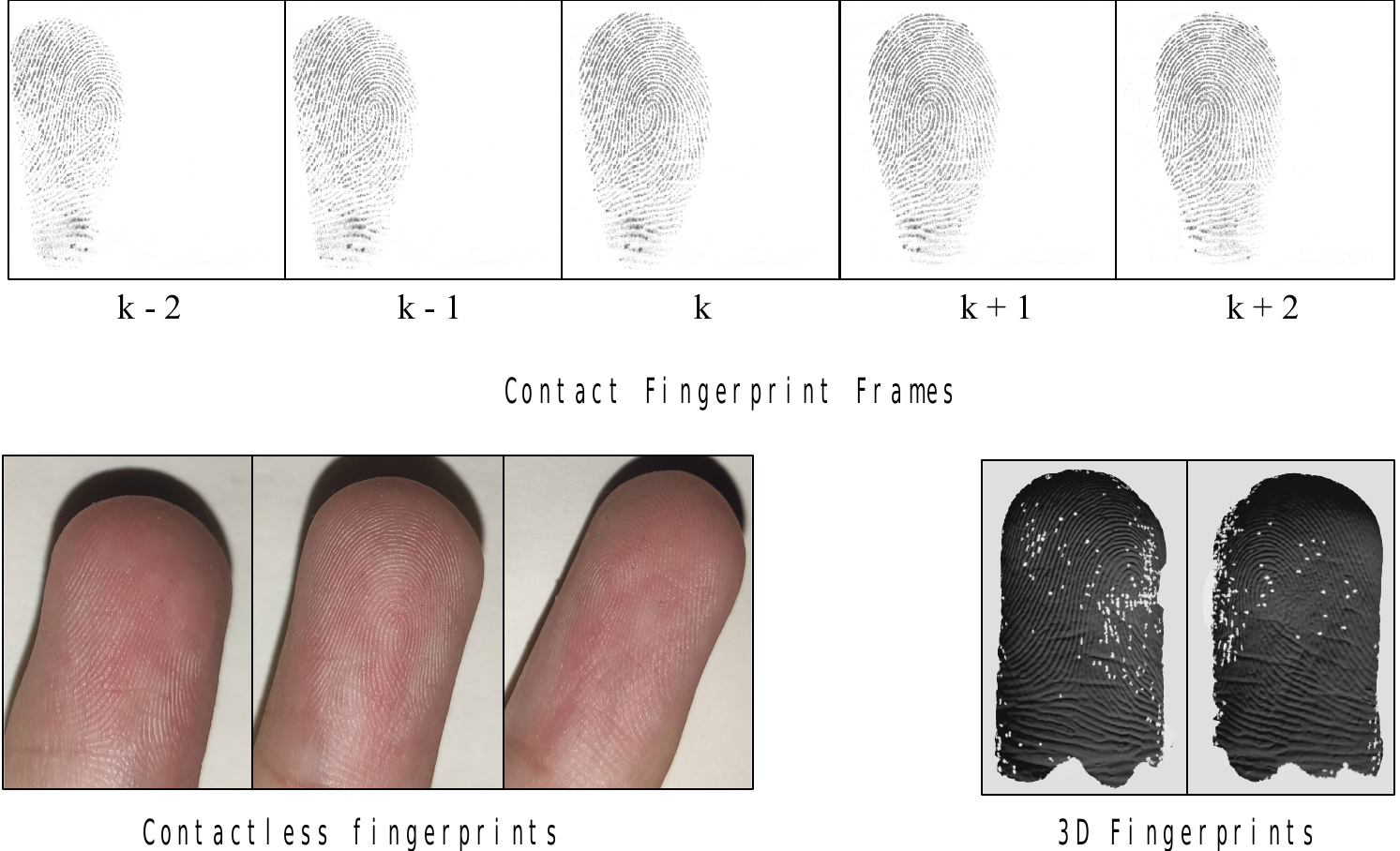}
\caption{Representative examples from the self-collected multimodal fingerprint database.}
\label{fig:data_examples}
\end{figure}

\begin{figure}[!t]
\centering
\subcaptionbox{3D scanner\label{fig:device3d}}{\includegraphics[width=0.45\linewidth]{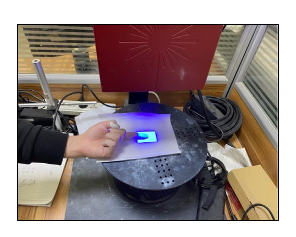}}
\hfill
\subcaptionbox{Contact sensor\label{fig:device2d}}{\includegraphics[width=0.45\linewidth]{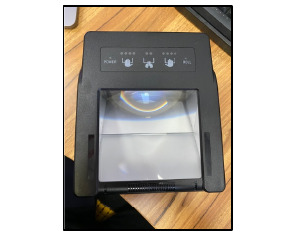}}
\caption{Fingerprint acquisition devices used in the experiments.}
\label{fig:devices}
\end{figure}

\begin{figure}[!t]
\centering
\subcaptionbox{Contact 2D}{\includegraphics[width=0.3\linewidth]{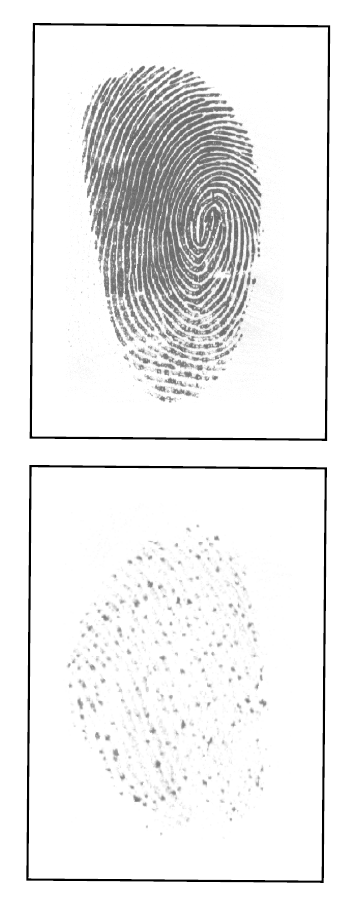}}
\hfill
\subcaptionbox{Contactless 2D}{\includegraphics[width=0.3\linewidth]{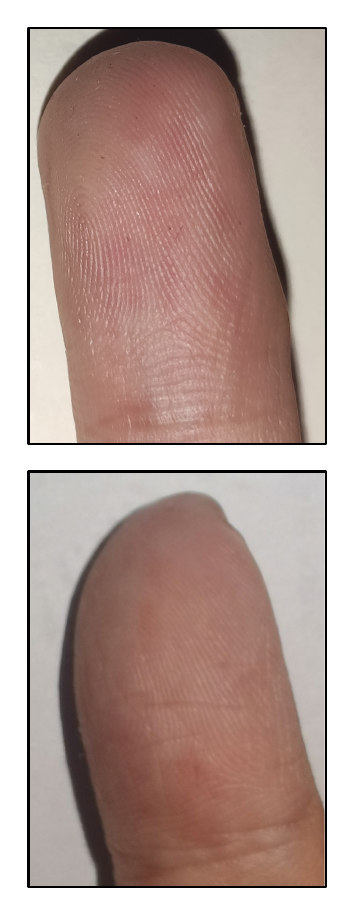}}
\hfill
\subcaptionbox{3D}{\includegraphics[width=0.3\linewidth]{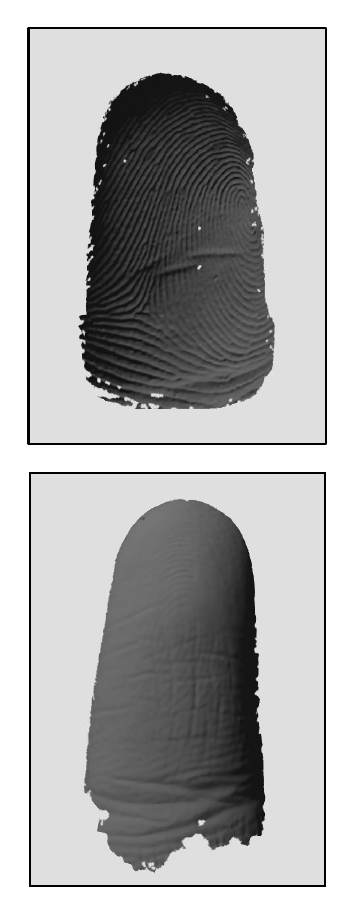}}
\caption{Representative high-quality and low-quality samples in the collected dataset.}
\label{fig:data_quality}
\end{figure}

Representative acquisition devices are shown in Fig.~\ref{fig:devices}. Representative quality variations of the three modalities are shown in Fig.~\ref{fig:data_quality}. In particular, contactless 2D fingerprints are highly sensitive to image blur and illumination, while 3D scans may lose ridge detail when the ridge height is too shallow.
In our acquisition experience, contact-based 2D fingerprints are usually the most stable modality, but low finger pressure may yield shallow ridges and insufficient contact area. Contactless 2D fingerprints are the most sensitive to camera parameters and lighting conditions, and many images become unusable because ridge-valley patterns are not clearly visible. The 3D scans usually preserve global finger shape well, but weak ridge relief can make the ridge texture difficult to recover from the surface geometry.

\subsection{Evaluation Protocol}
We evaluate the proposed system from four perspectives.
\begin{itemize}
\item \textit{3D fusion accuracy}: measured by the mean depth discrepancy between aligned point clouds after orthographic projection.
\item \textit{Pose-normalization robustness}: measured by the difference between the injected pose perturbation and the re-estimated pose after re-normalization.
\item \textit{Contactless 2D--3D registration accuracy}: measured by the projection error of aligned 3D minutiae onto the contactless image.
\item \textit{Contact-based 2D--3D compatibility}: measured by genuine matching scores computed by a commercial matcher \cite{verifinger}.
\end{itemize}

Because many fingerprint systems are closed-source and existing public datasets are not directly comparable to our multimodal acquisition setup, the experiments emphasize within-dataset validation of each stage instead of claiming a definitive benchmark comparison.
More concretely, we do not present a broad algorithmic baseline comparison for three reasons. First, many fingerprint systems relevant to this problem are closed-source or deeply integrated into commercial software stacks, making step-wise evaluation difficult. Second, existing methods are usually tested on datasets with different sensing modalities and incomplete correspondence to our setting, so direct comparison would be of limited value. Third, 3D fingerprint preprocessing and cross-modal registration still lack a widely accepted unified benchmark protocol.

\begin{table}[!t]
\centering
\small
\caption{Summary of main quantitative observations}
\begin{tabular}{>{\raggedright\arraybackslash}p{0.50\linewidth}c}
\toprule
Metric & Observation\\
\midrule
3D fusion error & concentrated around 0.09 mm\\
Representative ridge width & about 0.2 mm\\
Contactless 2D--3D projection error & around 6 pixels at 96 dpi\\
Pose error median & around $2^\circ$\\
Most pose errors & below $5^\circ$\\
Pose-aware pipeline runtime & about 14 s\\
\bottomrule
\end{tabular}
\label{tab:summary_metrics}
\end{table}

\subsection{3D Fusion Accuracy}
\begin{figure}[!t]
\centering
\includegraphics[width=0.9\linewidth]{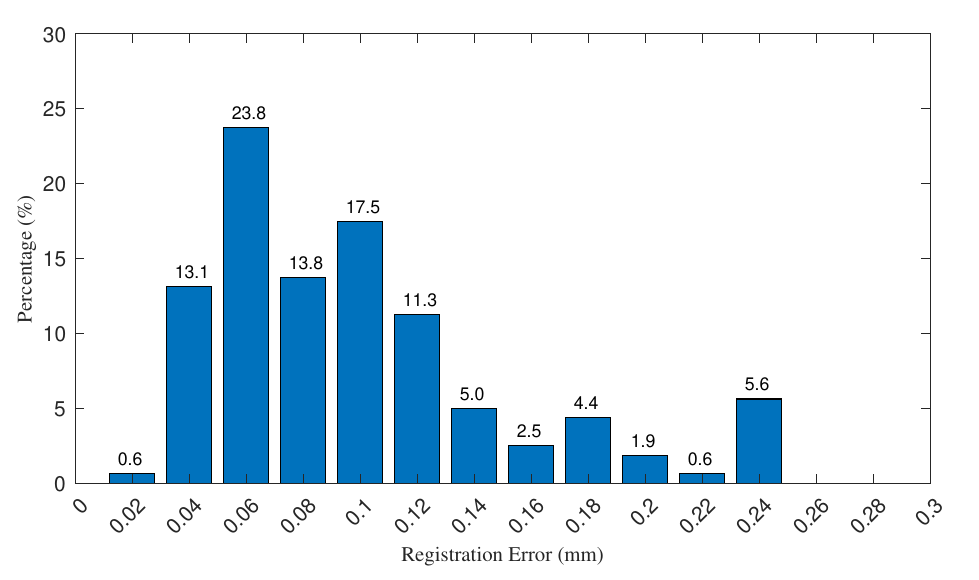}
\caption{Histogram of point-cloud registration errors for 3D fingerprint fusion.}
\label{fig:registration_err}
\end{figure}

To evaluate 3D fusion, we measure the average depth discrepancy between two aligned point clouds in their overlap region after projection. The histogram of registration errors is shown in Fig.~\ref{fig:registration_err}. Most errors are concentrated around 0.09~mm, whereas the ridge width in our data is approximately 0.2~mm. This indicates that the alignment reaches ridge-level precision and is sufficiently accurate for preserving ridge structure during fusion.

\begin{figure}[!t]
\centering
\subcaptionbox{Small overlap region}{\includegraphics[width=0.45\linewidth]{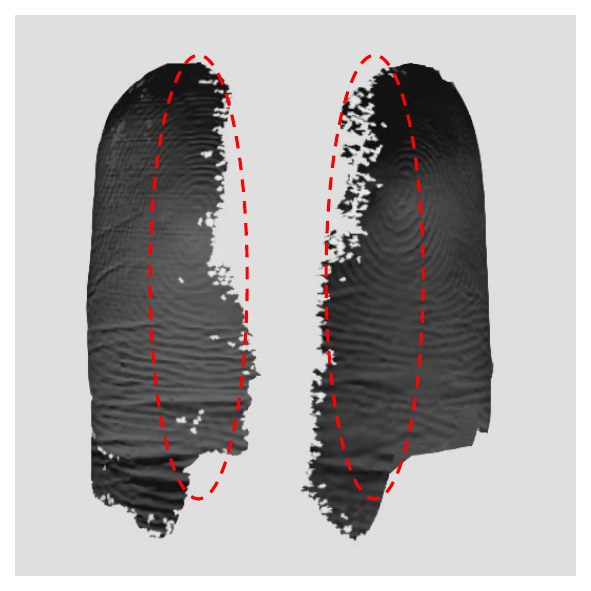}}
\hfill
\subcaptionbox{Low-quality scan}{\includegraphics[width=0.45\linewidth]{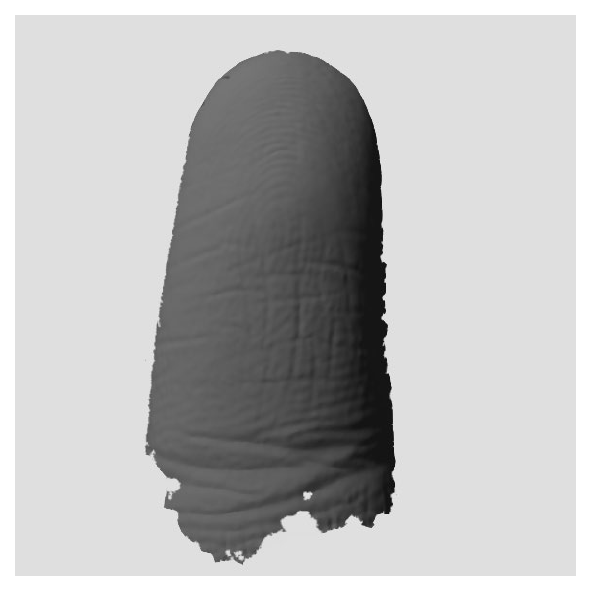}}
\caption{Representative failure cases for 3D point-cloud registration.}
\label{fig:bad_register}
\end{figure}

Failure cases are mainly caused by two factors. First, some scan pairs have only a small overlap region, which weakens both coarse correspondence estimation and rigid refinement. Second, some scans contain holes or weak ridge texture, which degrades point-cloud consistency. Representative difficult cases are shown in Fig.~\ref{fig:bad_register}.

\subsection{Pose-Normalization Robustness}
We evaluate pose normalization by first normalizing a finger point cloud, then rotating it by a known angle, and finally running the normalization procedure again. Let $\boldsymbol{\beta}_r$ be the injected rotation and $\boldsymbol{\beta}_e$ be the re-estimated pose. The error is measured as $|\boldsymbol{\beta}_e-\boldsymbol{\beta}_r|$.

\begin{figure*}[!t]
\centering
\subcaptionbox{}{\includegraphics[width=0.23\linewidth]{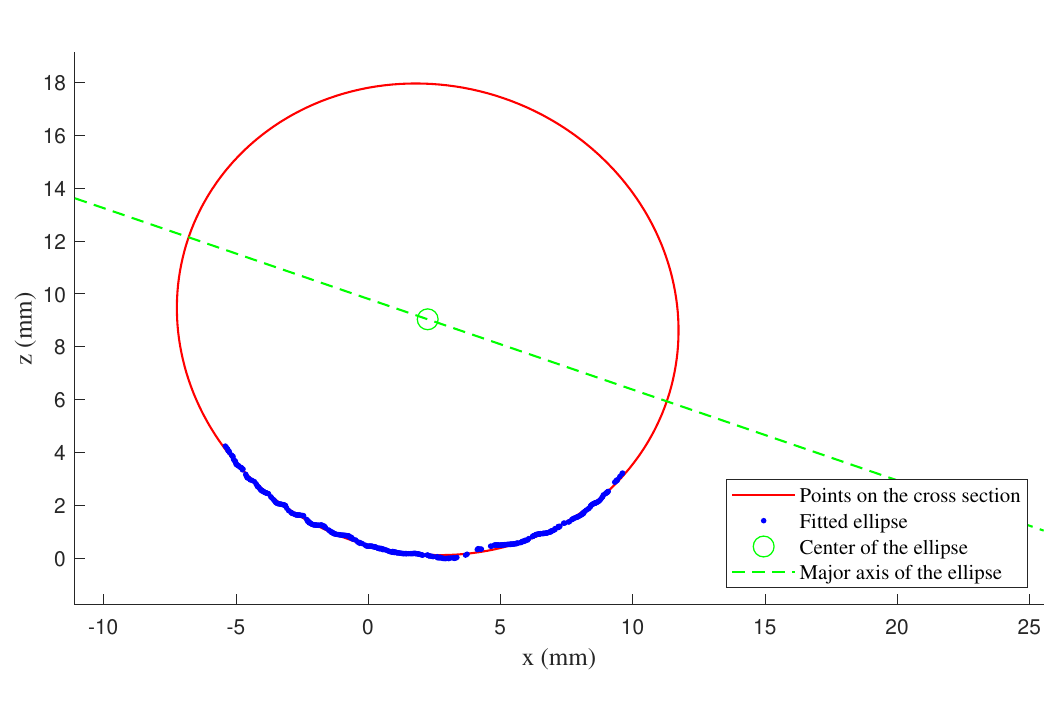}}
\hfill
\subcaptionbox{}{\includegraphics[width=0.23\linewidth]{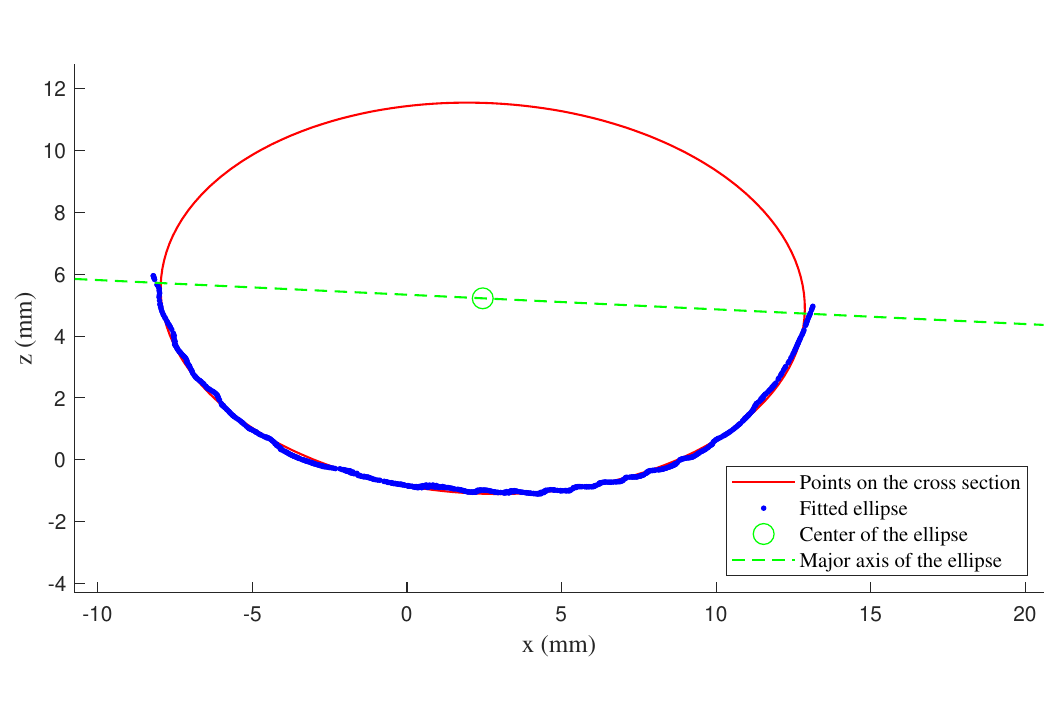}}
\hfill
\subcaptionbox{}{\includegraphics[width=0.23\linewidth]{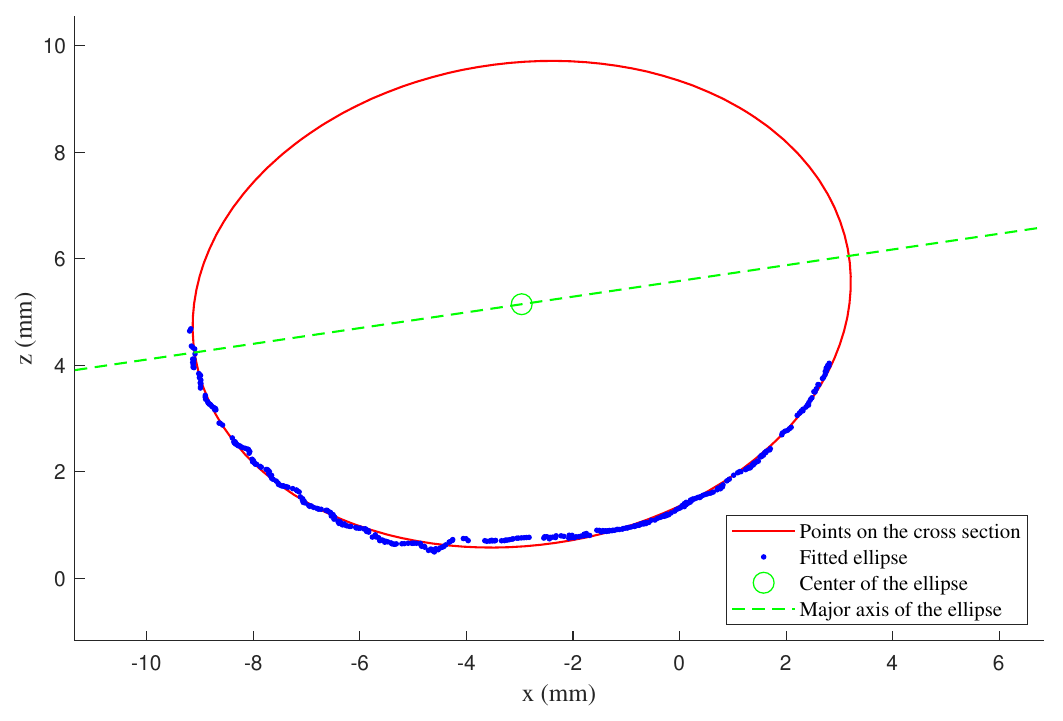}}
\hfill
\subcaptionbox{}{\includegraphics[width=0.23\linewidth]{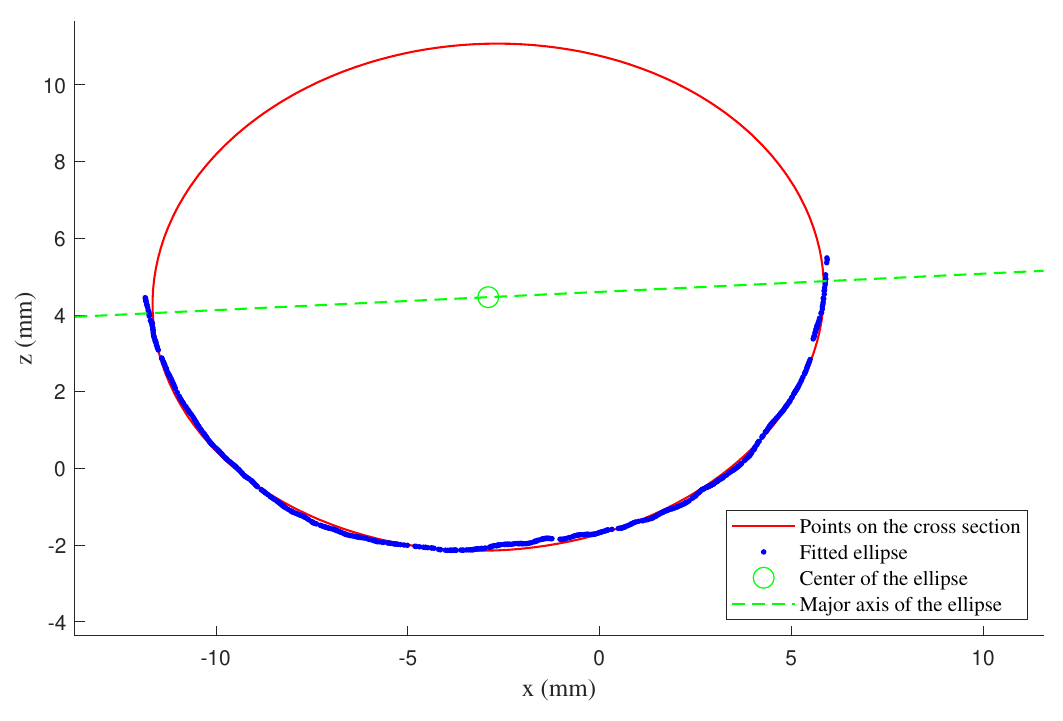}}\\
\subcaptionbox{}{\includegraphics[width=0.23\linewidth]{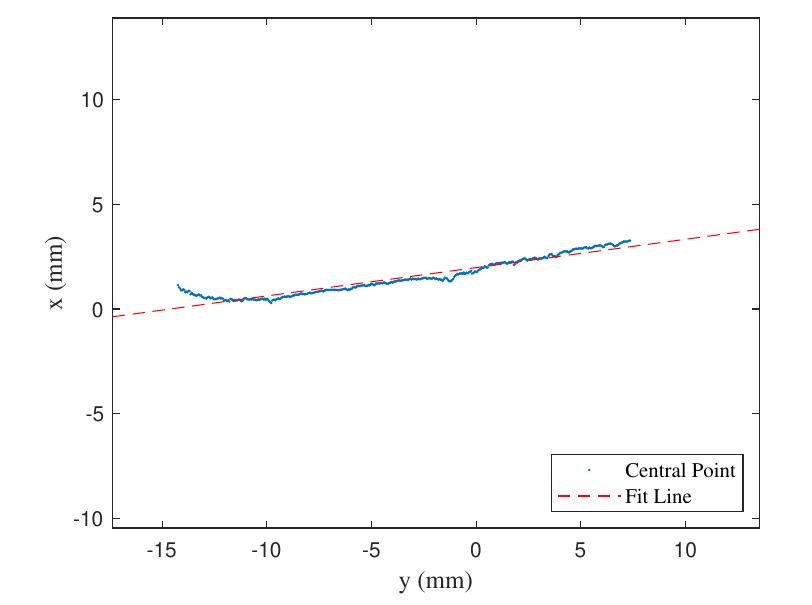}}
\hfill
\subcaptionbox{}{\includegraphics[width=0.23\linewidth]{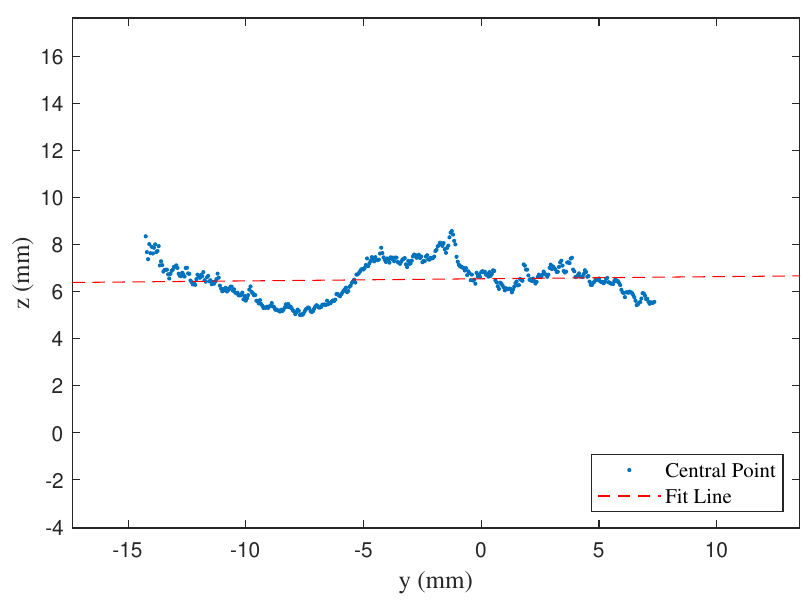}}
\hfill
\subcaptionbox{}{\includegraphics[width=0.23\linewidth]{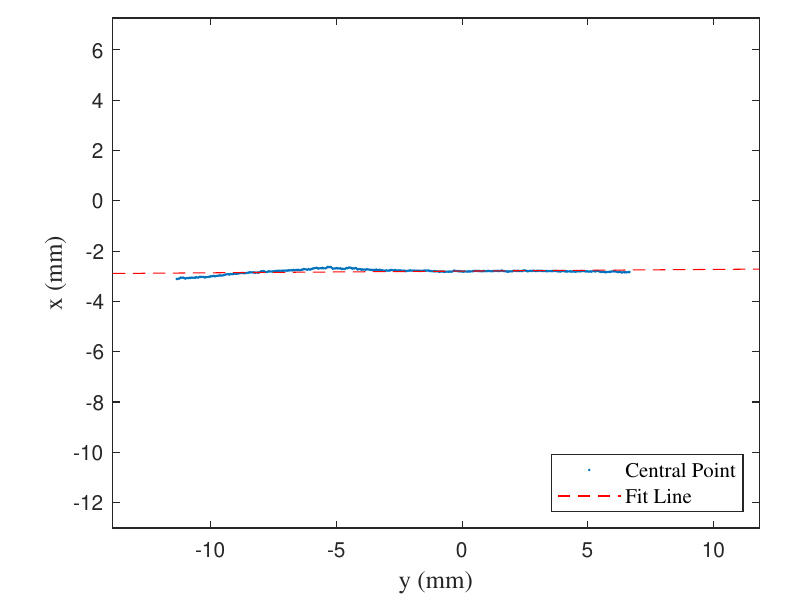}}
\hfill
\subcaptionbox{}{\includegraphics[width=0.23\linewidth]{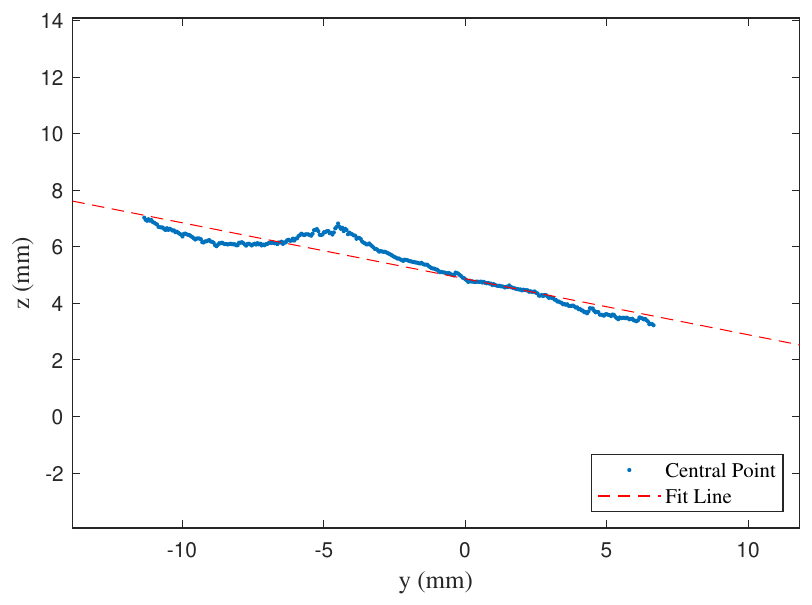}}
\caption{Ellipse fitting and center-axis estimation on representative thumb and index-finger point clouds.}
\label{fig:fit_ellipse}
\end{figure*}

Qualitative ellipse-fitting results on representative thumb and index-finger samples are shown in Fig.~\ref{fig:fit_ellipse}. The fitted ellipses closely follow the sampled cross-sections, and the estimated center points are approximately linear in both principal cross-sectional views, supporting the geometric assumptions used in the method.

\begin{figure}[!t]
\centering
\includegraphics[width=\linewidth]{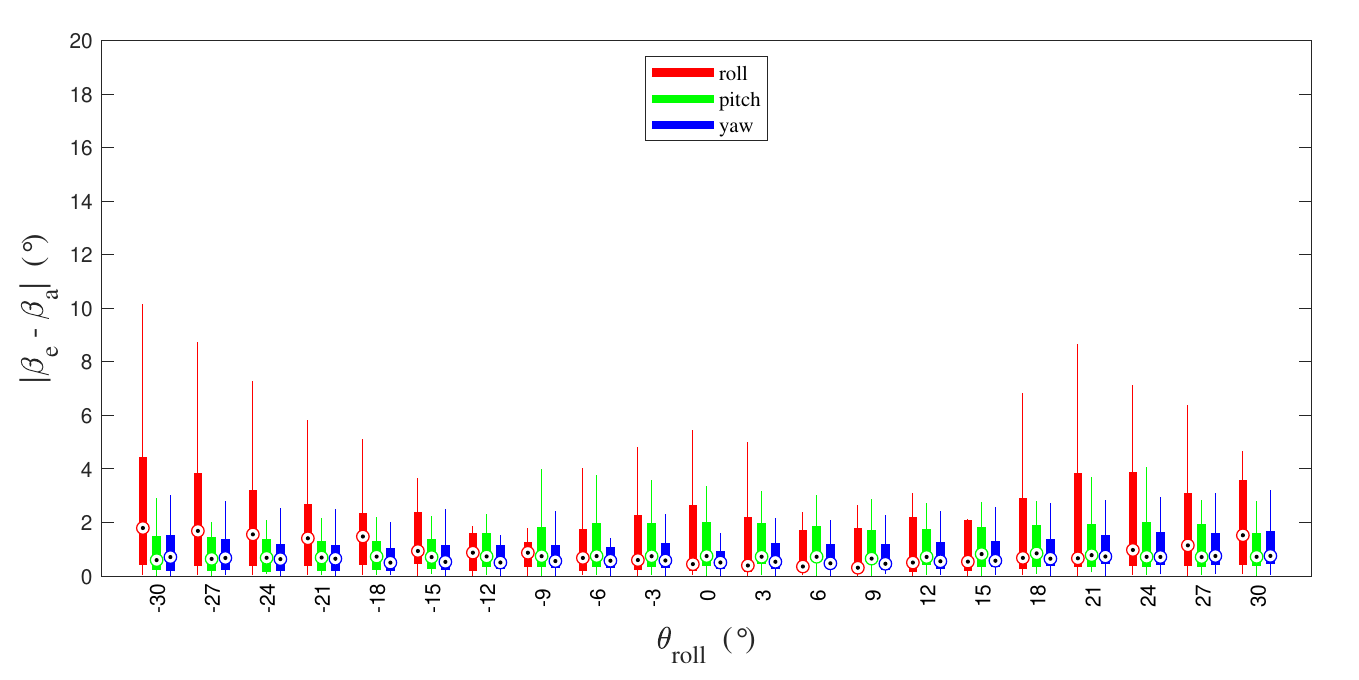}
\includegraphics[width=\linewidth]{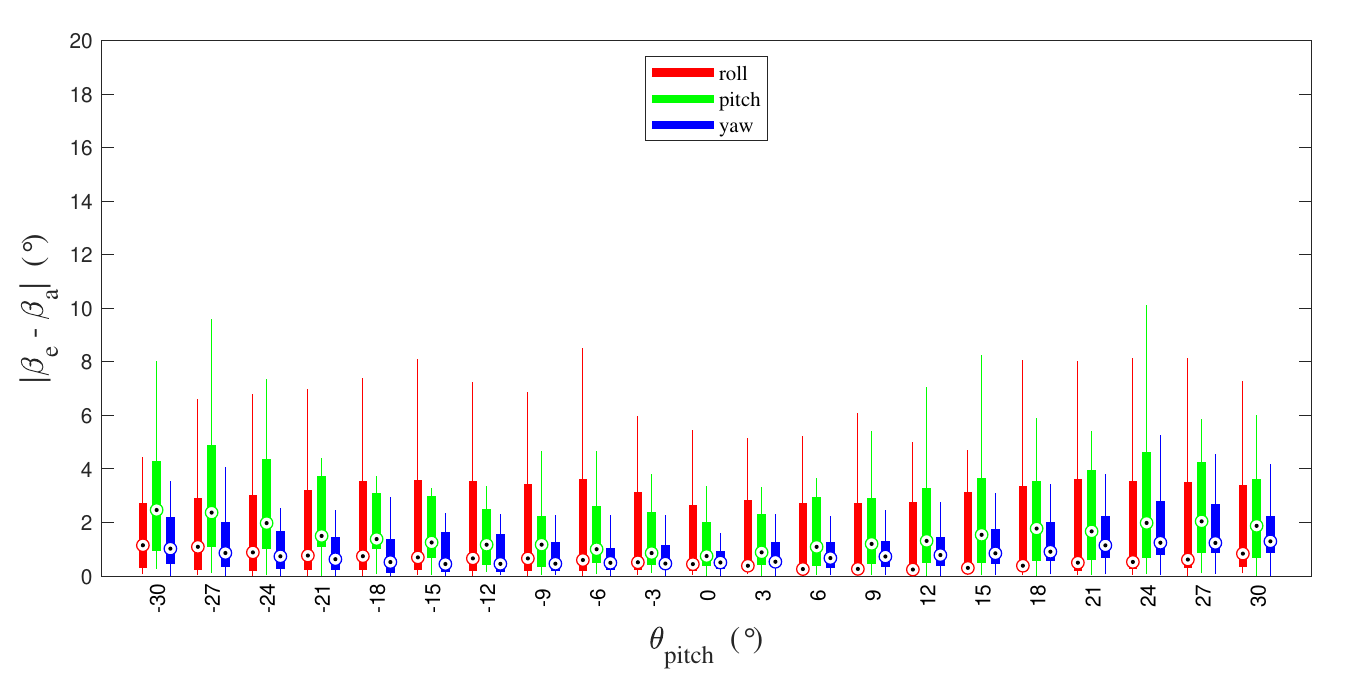}
\includegraphics[width=\linewidth]{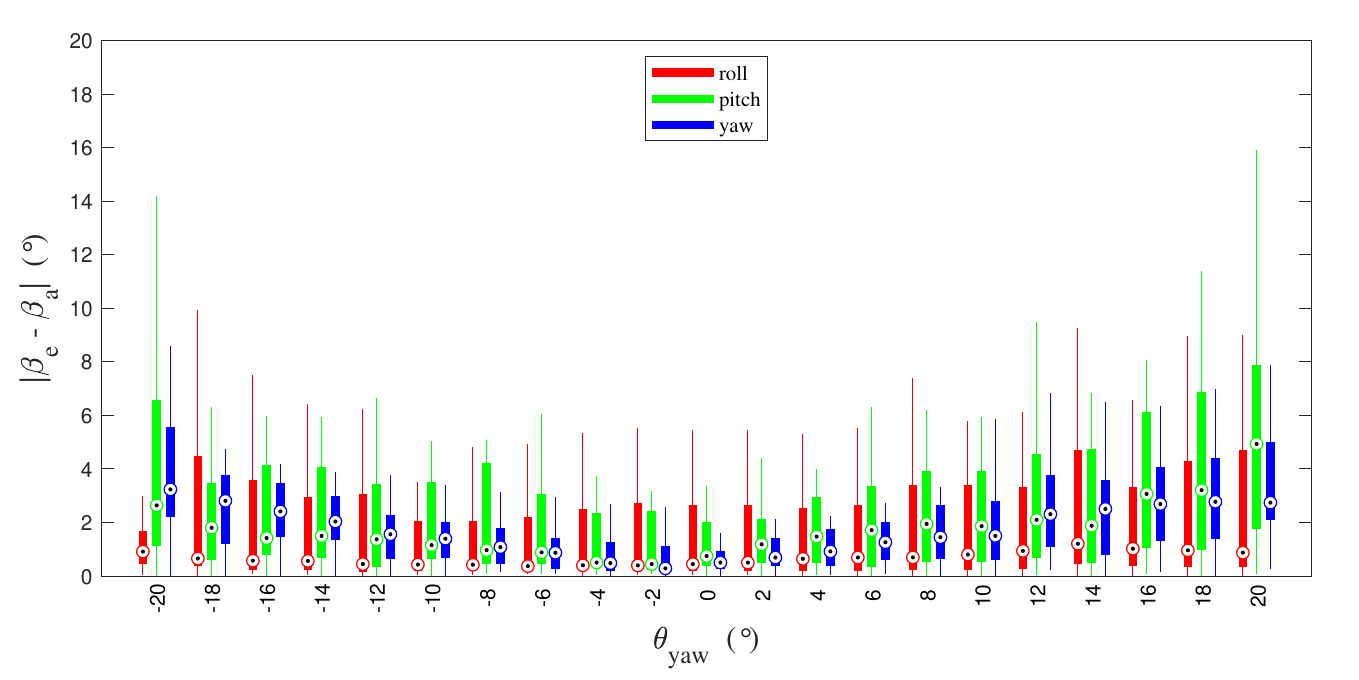}
\caption{Representative thumb pose-normalization errors for roll, pitch, and yaw.}
\label{fig:pose_err}
\end{figure}

Quantitatively, the median errors of pitch, roll, and yaw are around $2^\circ$, and most errors remain below $5^\circ$. Representative box plots are shown in Fig.~\ref{fig:pose_err}. Pitch and yaw are generally more stable than roll, which is consistent with the fact that the cross-sectional ellipse becomes less distinctive when the finger shape is nearly circular.

\begin{figure}[!t]
\centering
\includegraphics[width=\linewidth]{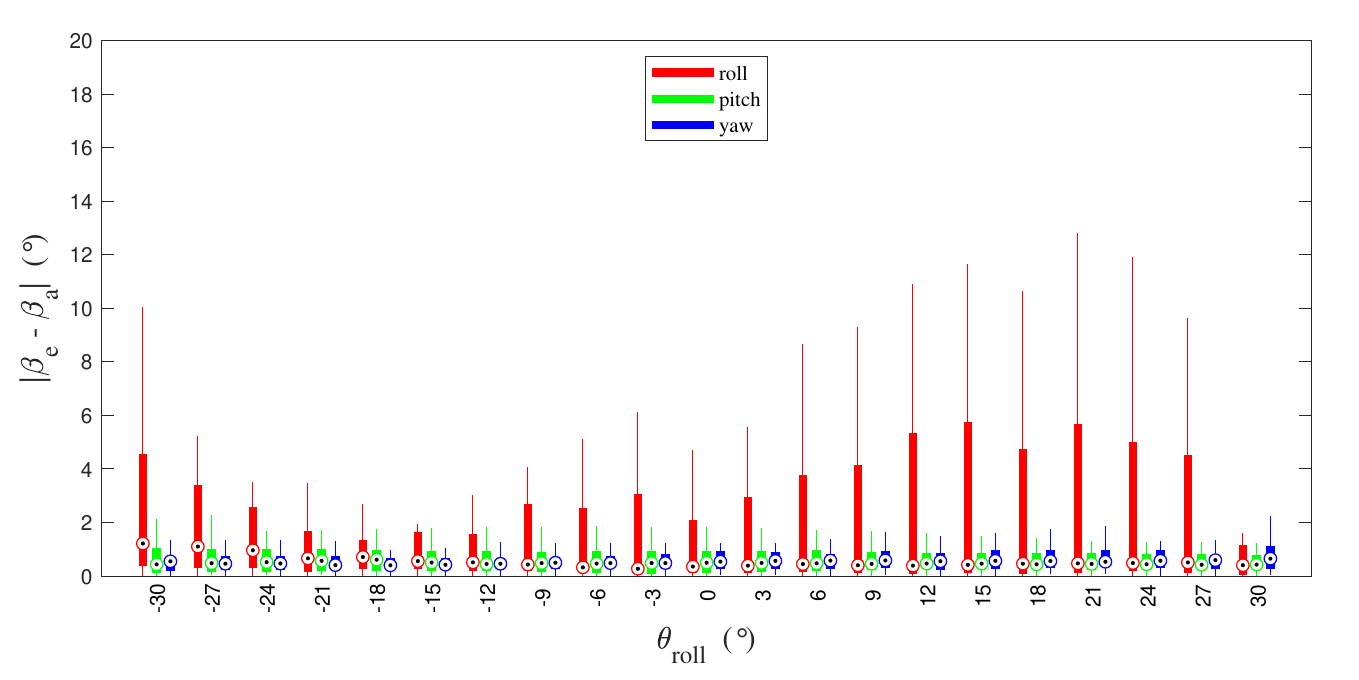}
\includegraphics[width=\linewidth]{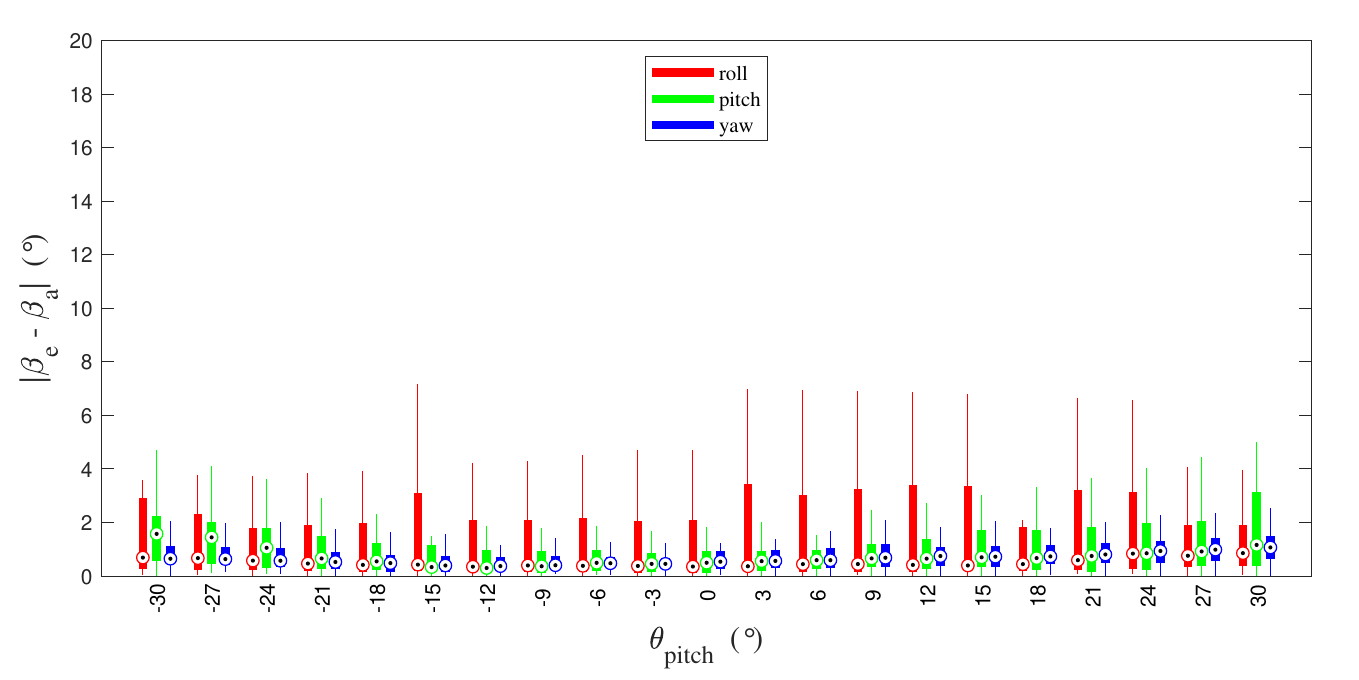}
\includegraphics[width=\linewidth]{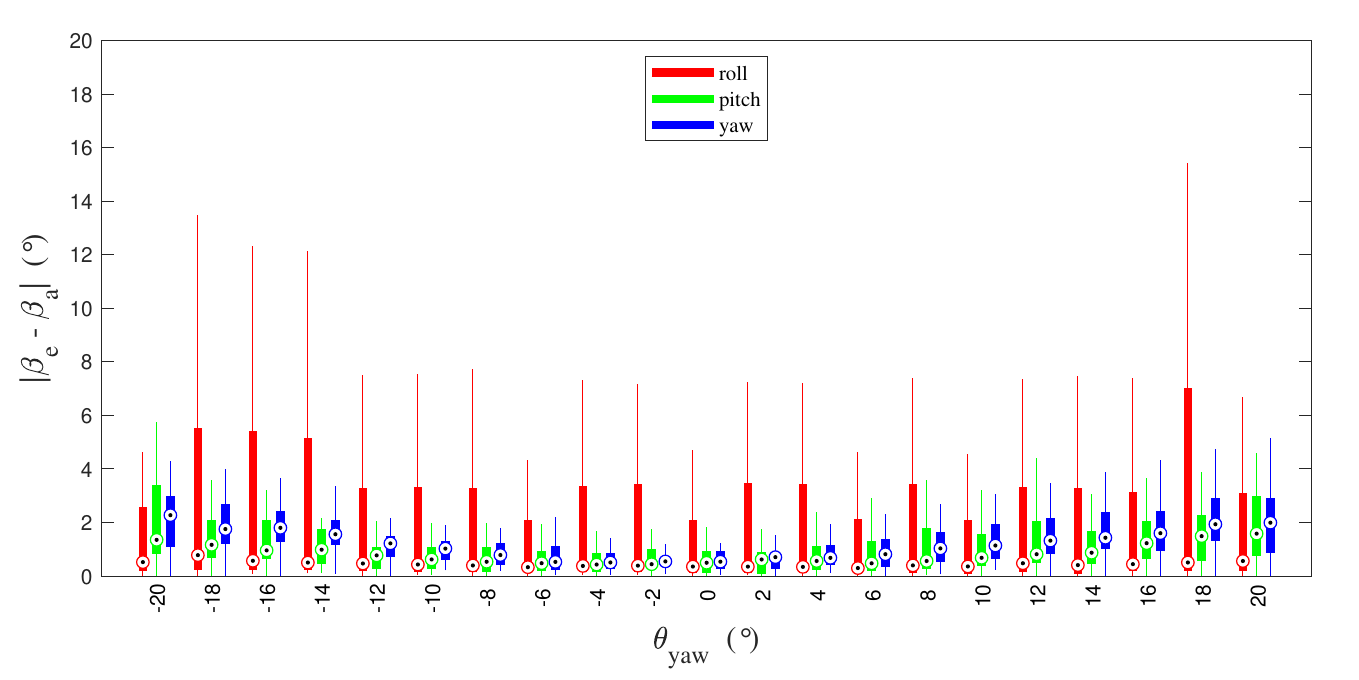}
\caption{Representative index-finger pose-normalization errors for roll, pitch, and yaw.}
\label{fig:index_pose_err}
\end{figure}

The result is consistent with the geometric interpretation of the method. Pitch and yaw are mainly constrained by the fitted center axis and are therefore relatively stable, whereas roll depends more strongly on the anisotropy of each cross-sectional ellipse and becomes harder to estimate when the finger section approaches a circular shape.
This trend is more pronounced for the index finger than for the thumb. The thumb usually has a thicker and flatter cross-section, so its ellipse parameters are more distinctive. In contrast, the index finger is closer to a circular cross-section, which makes major-axis estimation less reliable and explains the larger roll error observed in our experiments.

\subsection{Contactless 2D--3D Registration Accuracy}
\begin{figure}[!t]
\centering
\subcaptionbox{}{\includegraphics[width=0.45\linewidth]{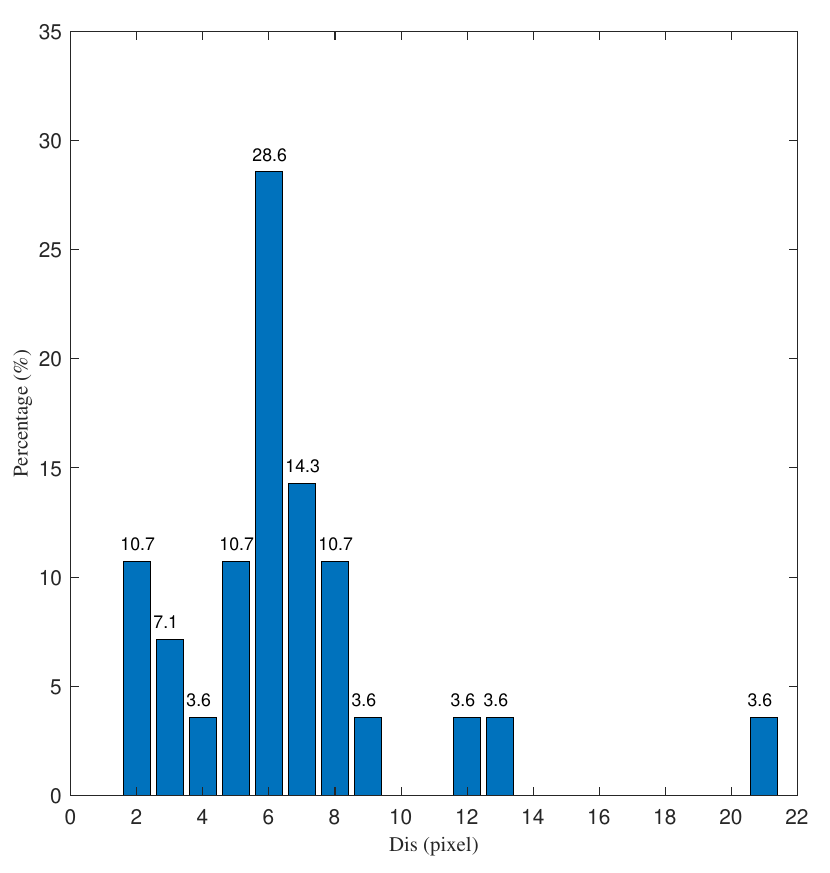}}
\hfill
\subcaptionbox{}{\includegraphics[width=0.45\linewidth]{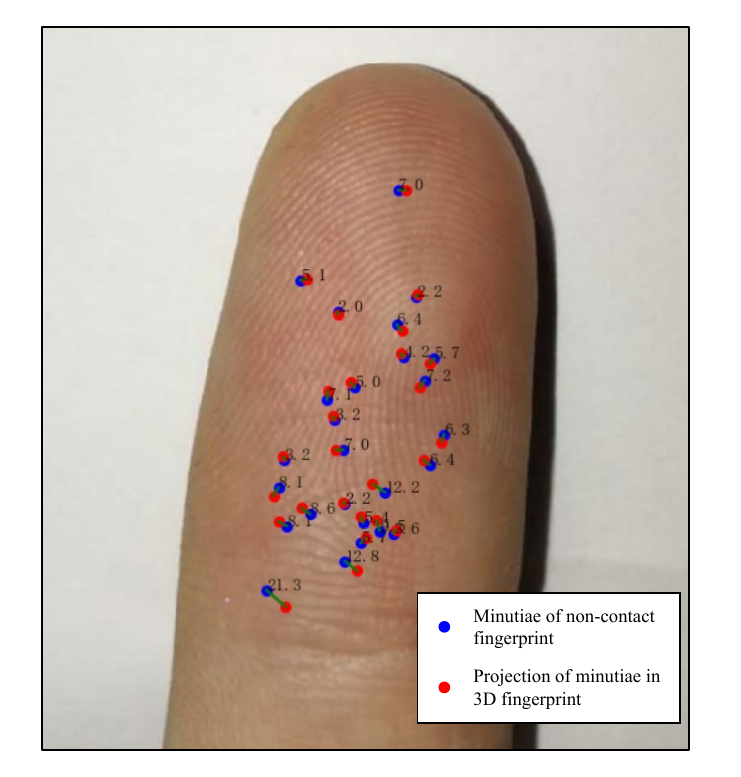}}
\caption{Projection error distribution and example for contactless 2D--3D registration.}
\label{fig:camera_index_err}
\end{figure}

For contactless 2D to 3D registration, the matched 2D--3D minutiae are used to estimate a camera-like projection model. The projection errors for representative contactless samples are shown in Fig.~\ref{fig:camera_index_err}. On 96-dpi contactless images, the projection errors are concentrated around 6 pixels, which is approximately on the scale of a single ridge width. This confirms that when reliable minutia correspondences are available, the proposed projection model yields ridge-level cross-modal alignment accuracy.

The main limitation of this branch is not the geometric model itself, but the difficulty of extracting enough reliable minutiae from contactless 2D fingerprints under unconstrained imaging conditions. When the 2D feature quality is sufficient, the alignment is stable and accurate.
For this reason, the current evaluation of the contactless 2D--3D branch is representative-case based rather than large-scale statistical benchmarking. The available samples still demonstrate that, under sufficiently reliable minutia correspondences, the proposed camera-model-based alignment can achieve ridge-scale geometric consistency.

\subsection{Contact-Based 2D--3D Registration Performance}
The most practically relevant experiment concerns the compatibility between 3D fingerprints and contact-based 2D fingerprints. We compare two strategies: direct generic unwrapping and the proposed pose-aware unwrapping. Matching is performed with a commercial fingerprint matcher, and higher genuine matching scores indicate better compatibility.

\begin{figure}[!t]
\centering
\subcaptionbox{\label{plot_register}}{\includegraphics[width=0.45\linewidth]{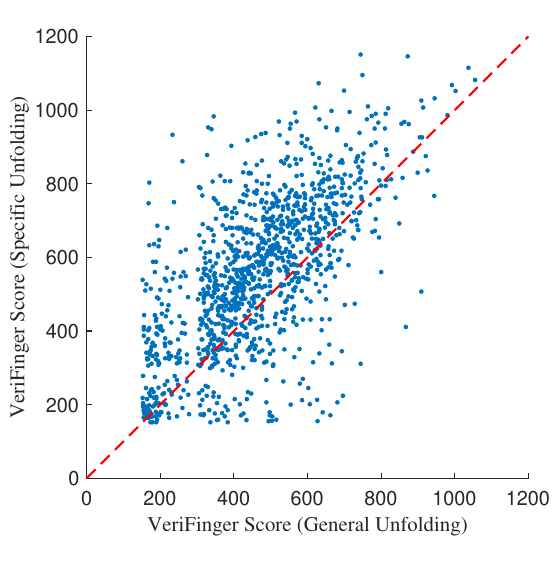}}
\hfill
\subcaptionbox{\label{hist_register}}{\includegraphics[width=0.45\linewidth]{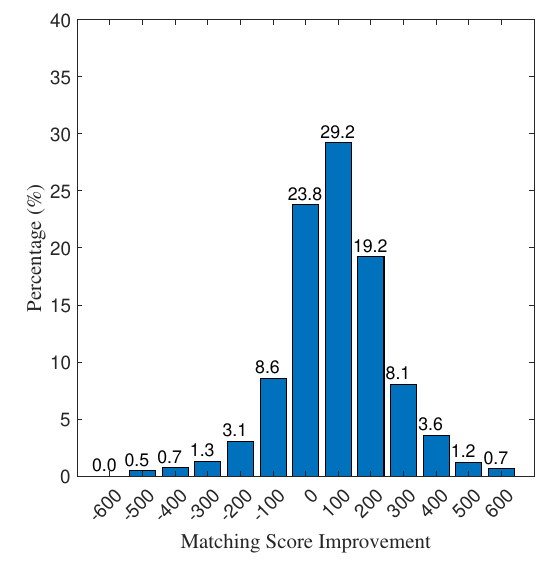}}
\caption{Comparison of genuine matching scores between direct unwrapping and pose-aware unwrapping.}
\label{fig:regist}
\end{figure}

The score comparison is shown in Fig.~\ref{fig:regist}. Most points lie above the diagonal in the scatter plot, indicating that pose-aware unwrapping improves compatibility for most genuine pairs. The histogram of score improvement further shows that positive gains dominate. Failures mainly occur when the initial minutia matches between the generic unwrapped image and the target contact fingerprint are too sparse, which leads to inaccurate pose estimation.

This result is important from a practical deployment perspective. The proposed method does not require replacing the legacy matcher; instead, it reshapes the 3D gallery sample into a more query-compatible representation. The gain therefore comes from better geometric compatibility rather than from any change in the final matching engine.
In other words, the proposed method reduces the deformation that the matcher has to tolerate, instead of asking the matcher itself to absorb all cross-modal discrepancy. This design choice makes the method attractive for legacy deployments where the 2D matcher is fixed.

\begin{figure*}[!t]
\centering
\includegraphics[width=0.98\linewidth]{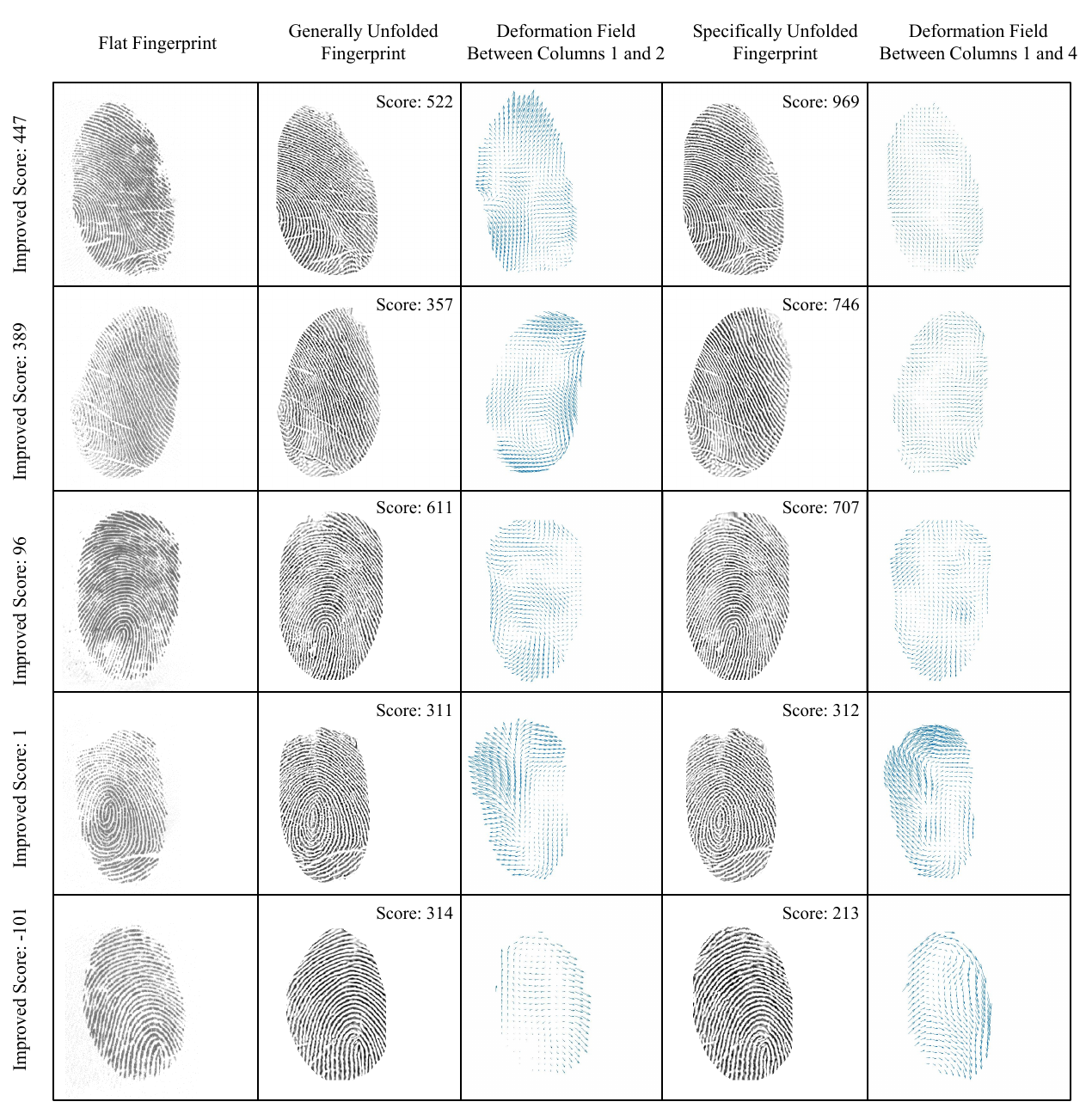}
\caption{Representative deformation-field comparisons between direct unwrapping and pose-aware unwrapping.}
\label{fig:deformation}
\end{figure*}

To better explain the score improvement, we estimate deformation fields between unwrapped 3D fingerprints and their corresponding contact-based 2D fingerprints using matched minutiae and thin-plate spline interpolation. Representative examples are shown in Fig.~\ref{fig:deformation}. Larger score gains are consistently associated with visibly reduced deformation fields, supporting the claim that pose-aware unwrapping reduces pose-induced cross-modal distortion.

We also report the runtime of the pose-aware unwrapping pipeline. The average execution times of the main processing stages are listed in Table~\ref{tab:speed}. The full pipeline takes approximately 14~s on a 2.50~GHz CPU. The additional cost over generic unwrapping mainly comes from 3D pose estimation, rigid transformation, and the second unwrapping step.

\begin{table}[!t]
\centering
\small
\caption{Average runtime of the pose-aware 3D-to-2D unwrapping pipeline}
\begin{tabular}{>{\raggedright\arraybackslash}p{0.58\linewidth}c}
\toprule
Step & Time (s)\\
\midrule
General unwrapping & 3.08\\
Feature extraction & 2.17\\
3D pose estimation & 2.23\\
3D rigid transformation & 1.44\\
Pose-aware unwrapping & 3.12\\
Final 2D alignment and cropping & 2.32\\
\bottomrule
\end{tabular}
\label{tab:speed}
\end{table}

\subsection{Contactless-to-Contact Registration via 3D Bridging}
\begin{figure}[!t]
\centering
\subcaptionbox{\label{2D_con_uncon1}}{\includegraphics[width=0.32\linewidth]{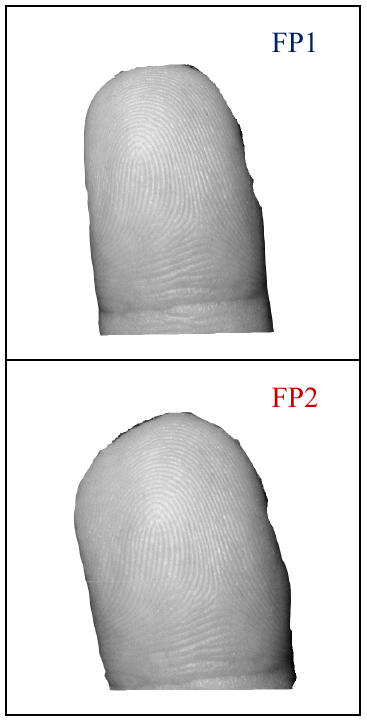}}
\hfill
\subcaptionbox{\label{2D_con_uncon2}}{\includegraphics[width=0.63\linewidth]{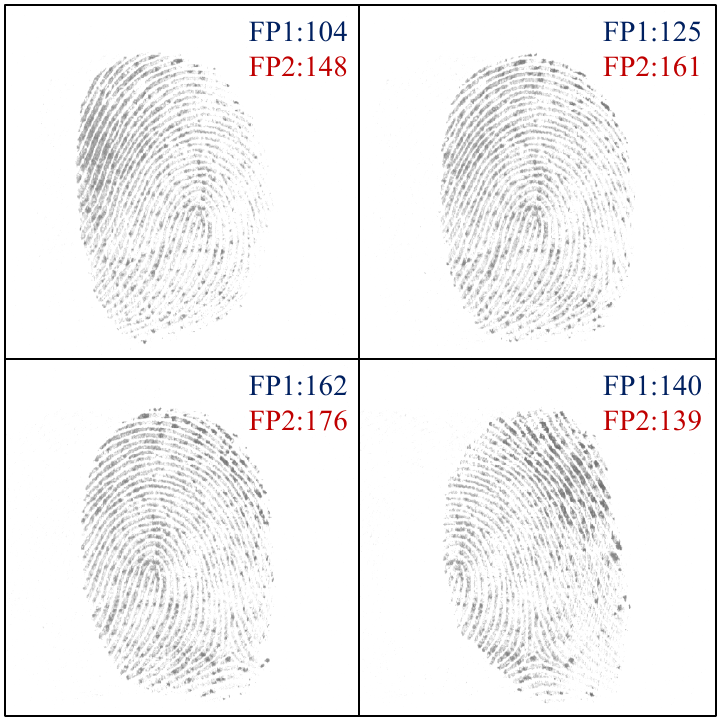}}
\caption{Representative improvement of contactless-to-contact matching by using 3D-guided unwrapping as an intermediate bridge.}
\label{fig:2D_con_uncon}
\end{figure}

Once the contactless 2D fingerprint is aligned with the 3D point cloud, the same 3D geometry can be used to unwrap the contactless image into a form that is more compatible with a contact-based fingerprint. Representative examples are shown in Fig.~\ref{fig:2D_con_uncon}. Compared with the original contactless image, the geometry-guided unwrapped image yields higher matching scores against contact-based fingerprints, especially when the original contactless view contains stronger perspective distortion.

\subsection{Submission-Oriented Interpretation}
From a submission-oriented perspective, the experiments support three main takeaways. First, the preprocessing stages are sufficiently accurate to preserve ridge-scale geometry rather than merely coarse finger shape. Second, the contactless 2D--3D branch is geometrically sound but still limited by the quality of contactless feature extraction. Third, the clearest practical gain comes from pose-aware unwrapping for contact-based 2D matching, where geometric normalization directly improves compatibility with legacy matching engines. This last result is the strongest system-level argument for the proposed framework.

\subsection{Discussion}
The experiments support the main premise of this work: 3D geometry is valuable not only as an additional biometric cue, but also as an intermediate representation for reducing deformation across heterogeneous fingerprint modalities. The proposed framework performs reliably at the preprocessing and registration level, especially for improving compatibility between 3D fingerprints and contact-based 2D fingerprints. In addition, the fusion and pose-normalization results suggest that the gallery side can tolerate multi-view partial scans and larger pose variation before cross-modal matching.

Several limitations remain. First, the current study relies on a self-collected dataset and therefore does not yet provide a broad benchmark against strong public baselines. Second, the accuracy of the contactless branch depends heavily on the quality of minutia extraction from contactless images. Third, the current contact model mainly addresses pose-induced deformation; pressure variation, local slipping, and sensor-specific elastic effects are not explicitly modeled. These limitations point to the most important directions for a future journal-strength revision.

\subsection{Future Directions}
An especially promising direction is to move from geometry-aware preprocessing to geometry-aware rendering. In the current framework, the 3D point cloud mainly serves as an intermediate geometric representation for visualization, unwrapping, fusion, and pose normalization. A natural next step is to use the recovered 3D structure to directly render fingerprint observations under richer pose and sensing conditions. Such a rendering-oriented view would allow the 3D gallery sample to generate a family of pose-conditioned 2D fingerprints rather than only a single canonical unwrapped image, which could further improve interoperability with heterogeneous query images.

Recent progress in neural rendering and view synthesis suggests several possibilities. A fingerprint-specific rendering model could combine point clouds, surface normals, ridge-relief cues, and observed 2D fingerprint images to synthesize realistic fingerprints under novel viewpoints. In this setting, non-contact 2D images and contact-based 2D fingerprints would no longer be treated only as matching targets; they could also act as supervision for reconstructing or refining a richer 3D appearance model. Compared with classical unwrapping alone, such a representation may preserve both geometric consistency and modality-specific image formation effects, including perspective variation, local shading changes, and incomplete ridge visibility.

Another valuable direction is joint rendering from mixed observations. If multiple contactless views, contact-based impressions, and partial 3D point clouds are all available for the same finger, one could build a unified reconstruction-and-rendering model that explains these observations together. The resulting model could then render fingerprints at controlled yaw, pitch, and roll angles, or even approximate different contact conditions. From a system perspective, this would turn a limited set of enrolled samples into a much richer pose-diverse gallery and could reduce the need for exhaustive multi-pose acquisition during enrollment.

This rendering perspective is also attractive for data augmentation and benchmarking. A reliable 3D-guided renderer could generate cross-modal pairs with known pose and deformation relationships, making it easier to study registration accuracy under controlled conditions. It could further support future benchmarks in which 3D fingerprints are used not only as biometric samples, but also as generative geometric priors for synthesizing realistic contactless and contact-based fingerprints. In that sense, the current work can be viewed as a preprocessing-oriented first step toward richer 3D-driven fingerprint rendering and recognition systems.

\section{Conclusion}
This paper presented a unified framework for 3D fingerprint preprocessing and cross-modal registration. By combining local-geometry-based visualization, nonparametric unwrapping, point-cloud fusion, ellipse-based pose normalization, and pose-aware cross-modal alignment, the framework improves the compatibility of 3D fingerprints with both contactless and contact-based 2D fingerprints while preserving compatibility with legacy 2D matching software. Experiments on a self-collected multimodal dataset demonstrated ridge-level 3D fusion accuracy, stable pose normalization, and consistent gains in cross-modal compatibility, especially for contact-based 2D--3D matching. Overall, the results suggest that 3D fingerprints are most valuable here as a geometric bridge that reduces pose- and shape-induced mismatch across modalities rather than as an isolated sensing modality. Future work will focus on larger-scale evaluation, stronger baseline comparisons, more robust contactless feature extraction, and richer modeling of contact-induced elastic deformation.

\clearpage
\bibliographystyle{IEEEtran}
\bibliography{refs}

\end{document}